\documentclass[accepted]{uai2026} 
                        
\usepackage[american]{babel}

\usepackage{natbib} 
\usepackage{mathtools} 
\usepackage{nicefrac}
\usepackage{booktabs} 
\usepackage{tikz} 
\usepackage{multirow}
\usepackage{amsfonts, amsmath, amssymb, amsthm, thmtools}
\usepackage{subcaption}
\usepackage{hyperref}
\usepackage{tabularx}

\def\Q{{\mathbb Q}}

\DeclareMathOperator*{\esssup}{ess\,sup}
\DeclareMathOperator*{\essinf}{ess\,inf}

\newtheorem{theorem}{Theorem}

\newtheorem{proposition}{Proposition}

\renewcommand{\epsilon}{\varepsilon}

\title{Robustness Quantification for Discriminative Models:\\ A New Robustness Metric and its Application to Dynamic Classifier Selection}

\author[1,2,3]{\href{mailto:<rflassance@usp.br>?Subject=Your UAI 2026 paper}{Rodrigo~F.~L.~Lassance}{}}
\author[1]{\href{mailto:<jasper.debock@ugent.be>?Subject=Your UAI 2026 paper}{Jasper~De~Bock}{}}
\affil[1]{%
    Foundations Lab for imprecise probabilities (FLip)\\
    Ghent University\\
    Ghent, Belgium
}
\affil[2]{%
    Statistics Dept.\\
    Federal University of São Carlos\\
    São Carlos, São Paulo, Brazil
}
\affil[3]{%
    Institute of Mathematic and Computer Sciences\\
    University of São Paulo\\
    São Carlos, São Paulo, Brazil
  }
  
\begin{document}
\maketitle

\begin{abstract}
Among the different possible strategies for evaluating the reliability of individual predictions of classifiers, robustness quantification stands out as a method that evaluates how much uncertainty a classifier could cope with before changing its prediction. However, its applicability is more limited than some of its alternatives, since it requires the use of generative models and restricts the analyses either to specific model architectures or discrete features. In this work, we propose a new robustness metric applicable to any probabilistic discriminative classifier and any type of features. We demonstrate that this new metric is capable of distinguishing between reliable and unreliable predictions, and use this observation to develop new strategies for dynamic classifier selection.
\end{abstract}

\section{Introduction}\label{sec:intro}

Machine learning models possess great predictive capacity, but this capacity comes with a level of unreliability that is hard to assess. From a modeling perspective alone, the majority of the methods used to make high-stakes decisions lack transparency in what their true decision process is, making their users potentially subject to harmful consequences in the process \citep{oneil2016}. Although there have been attempts to make such black box models more interpretable \citep{molnar2025} or to switch to models that are inherently interpretable instead \citep{rudin2019}, these contributions remain insufficient due to reasons that go beyond the models themselves. After all, a model is only as good as the data that has been provided to it, with the inherent variability of the phenomena and the lack of all meaningful features leading to an unreliability source referred to as Aleatoric Uncertainty (AU). Moreover, when \textbf{(i)} the data is not sufficient for the model to differentiate pattern from noise to an acceptable degree or \textbf{(ii)} there is a discrepancy between the data being used to train the model and the context in which the model will be applied to, this also leads to another unreliability source: Epistemic Uncertainty (EU).

While completely removing these sources of uncertainty is impossible by definition, progress has been made in evaluating how reliable an individual prediction is in the field of uncertainty quantification. Many different approaches have already been explored \citep{hullermeier2021}, with the attempt of estimating and separating AU and EU among the most popular strategies. If successfully done, this allows decision makers to refrain from trusting the outputs of a model in specific circumstances, avoiding the more undesirable repercussions that could come. However, properly quantifying EU is rarely feasible from the data used for training alone, since it most often diverges from the true application that a practitioner has in mind. Aside from specific situations, some aspects of EU are fundamentally inaccessible in practice.

A promising alternative to directly estimating EU is using robustness quantification. Similarly to uncertainty quantification, earlier work \citep{detavernier2025rob, detavernier2025comb} has demonstrated that robustness quantification can be used to assess the reliability of individual predictions. Unlike uncertainty metrics, which aim to quantify the amount of uncertainty that influences the decision, robustness metrics aim to quantify how uncertain one can be about the model while still being able to trust its decision,
thereby sidestepping the problem of estimating the amount of uncertainty there actually is. Both types of metrics tend to correlate with the accuracy of the predictions, and their relative performance varies with the context. Robustness metrics tend to perform best exactly in the contexts where uncertainty is difficult to estimate, such as in the presence of distribution shift, model misspecification, or small data regimes. 

Compared to other applications of the term ``robustness'' in machine learning, one of the differentiating aspects of robustness quantification is that it is an instance-based metric, i.e., it is an assessment of how reliable a specific individual prediction is based on the underlying model and the values of the features used as input. It is also different from the notion of adversarial robustness \citep{muhammad2022}, which focuses on perturbing the features and is mainly used in applications with image data. Robustness quantification, on the other hand, comes from a notion of perturbing the joint distributions of the trained model, which is more in line with ideas in imprecise probabilities \citep{augustin2014}.

Important downsides of robustness quantification, at this point, are that it can only be applied to generative models, is mainly based on epsilon-contamination, and is restricted to either specific model architectures \citep{debock2014,correia2019, correia2020rob} or to fully discrete sets of features \citep{detavernier2025rob}.
In this work, we improve on these current limitations by proposing a new robustness metric that is applicable to any probabilistic discriminative classifier (\autoref{sec:setup}). Instead of using epsilon-contamination, our metric is based on the Constant Odds Ratio (COR) perturbation, yielding a metric that is not restricted to discrete features. We start in Section~\ref{sec:setup} with an introduction to robustness quantification and explain how, in particular, a dissimilarity function can be used to define a robustness metric. \autoref{thm:rob} provides a closed-form expression for the COR perturbation case that we focus on and is followed by a list of practical considerations that motivate this particular choice of perturbation. Section~\ref{sec:experiments} goes on to test the performance of this new metric. In a first batch of experiments, we use Accuracy Rejection Curves to demonstrate that this metric correlates nicely with accuracy, that it does this better than an alternative competitor, and that it can do this for several model architectures (section \ref{sec:corr}). Our main application uses robustness to perform dynamic selection of classifiers, offering two strategies for choosing which model to use as a predictor given a set of features (section \ref{sec:ds}). We end in Section~\ref{sec:disc} with a discussion that highlights possible venues for future research.

\section{Robustness Quantification}\label{sec:setup}

Let $Y\in\mathcal{Y}$ be a discrete class variable and $X\in\mathcal{X}$ its vector of features. The features can be purely discrete, purely continuous or a mixture of both. Uncertainty about $(Y,X)$ is expressed by a probability measure $P$ on a suitable sigma algebra $\mathcal{A}$ of subsets of $\mathcal{Y}\times\mathcal{X}$ (e.g. the power set when $X$ is purely discrete or a product of a power set and Borel sigma algebra if $X$ purely continuous or mixed). We furthermore assume that $P$ is absolutely continuous w.r.t.\ an adequate base measure $\mu$ (e.g., the counting measure when $X$ is purely discrete or the product of a counting and Lebesgue measure when $X$ is purely continuous or mixed), which guarantees that $P$ has a density $p=\nicefrac{dP}{d\mu}$ w.r.t. this base measure $\mu$. We denote the set of all such probability measures by $\mathbb{P}(\mathcal{Y},\mathcal{X})$.

We consider a
classification problem where the goal is to predict the value of $Y$ given $x$ based on $P$ or, equivalently, based on $p$. The classifier that minimizes the 0-1 loss then predicts the most likely class given the features $x$, which is given by
\begin{align}
    g_p(x) :&= \arg\max_{y\in\mathcal{Y}} p(y|x) = \arg\max_{y\in\mathcal{Y}} \frac{p(y,x)}{p(x)}\notag\\
    &= \arg\max_{y\in\mathcal{Y}} p(y,x).
\end{align}

We are interested in quantifying the robustness of this prediction, by determining how stable it is with respect to divergences from the original measure $P$. To this end, following \cite{detavernier2025rob}, we consider parametrized perturbations $\mathcal{P}_\delta$ around $P$, with $\delta \in \mathbb{R}_{\geq0}$. We call the prediction $g_p(x)$ robust w.r.t. $\mathcal{P}_\delta$ if, for all $P' \in \mathcal{P}_\delta$, the prediction $g_{p'}(x)$ is the same as $g_p(x)$. The robustness $r(x)$ of $g_p(x)$ is then quantified as the largest $\delta$ such that $g_p(x)$ is robust w.r.t. $\mathcal{P}_\delta$.

Work on robustness quantification has so far mainly focused on perturbations $\mathcal{P}_\delta$ based on epsilon-contamination. One approach consists in applying epsilon-contamination directly to the global model $P$, resulting in perturbations of the form $\mathcal{P}_\epsilon=\{(1-\epsilon)P + \epsilon Q: Q \in \mathbb{P}(\mathcal{Y},\mathcal{X})\}$ with $\epsilon\in[0,1]$. This leads to simple closed form expressions for the robustness metric $r(x)$, but it is only meaningful if $X$ is discrete since it otherwise typically leads to robustness values of zero. Another approach consists in applying epsilon-contamination to the local parameters of specific parametric models such as Naive Bayes classifiers \citep{detavernier2025rob}, Sum-Product Networks \citep{correia2019} or Generative Forests \citep{correia2020rob}. This local approach remains meaningful for continuous or mixed features as well (at least for Sum-Product Networks and Generative Forests), but is only applicable to specific parametric models and can be computationally expensive. 

In this work, we study robustness quantification based on a dissimilarity function $d$ between probability measures. That is, we consider perturbations of the type 
\begin{equation}\mathcal{P}_\delta = \{Q \in \mathbb{P}(\mathcal{Y},\mathcal{X}): d(P,Q) \le \delta\}\label{eq:perturbation}
\end{equation}
with $\delta \in \mathbb{R}_{\geq0}$. In other words, the prediction is robust w.r.t. $\mathcal{P}_\delta$ if the prediction $g_q(x)$ is the same as $g_p(x)$ for all $Q \in \mathbb{P}(\mathcal{Y},\mathcal{X})$ such that $d(P,Q) \le \delta$, and the robustness $r(x)$ of the prediction $g_p(x)$ is the largest $\delta$ for which this is the case. To indicate the influence of the choice of $d$, we will write $r_d(x)$ instead of $r(x)$ when referring to the robustness metric based on a specific dissimilarity function $d$.

An interesting property of a robustness metric based on a dissimilarity function $d$ is presented in the proposition below and follows by definition.
\begin{proposition}
    \label{prop:pred_match}
    For any $Q \in \mathbb{P}(\mathcal{Y}, \mathcal{X})$ and $x \in \mathcal{X}$:
    $$
        d(P,Q) < r_d(x) \Longrightarrow g_p(x) = g_q(x).
    $$
\end{proposition}
Notwithstanding the possibility of the measure $P$ differing substantially from the true Data Generating Process (DGP) of the new data, \autoref{prop:pred_match} guarantees that their predictions will be the same for inputs whose robustness is high enough. Therefore, if the DGP provides reliable predictions, then so does $P$ when robustness is high.

 We focus in particular on the distance function $d_{\mathrm{COR}}$, defined for all $P, Q \in \mathbb{P}(\mathcal{Y},\mathcal{X})$ by
\begin{equation}\label{eq:dcor}
    d_{\mathrm{COR}}(P, Q) := 
    \sup_{\substack{A,B\in\mathcal{A}\\Q(A)>0,P(B)>0}}\left(1-\frac{P(A)Q(B)}{Q(A)P(B)}\right).
\end{equation}
The idea is that $P$ and $Q$ are similar if the odds $\nicefrac{P(A)}{P(B)}$ and $\nicefrac{Q(A)}{Q(B)}$ are similar for any two events $A$ and $B$.
Similarly to epsilon-contamination, the perturbation induced by Equation~\eqref{eq:dcor} is related to imprecise probabilities. More specifically, \cite{montes2020} shows that the perturbations that correspond to this distance function are Constant Odds Ratio (COR) models~\citep[][Section 4.7.2]{augustin2014}, which can also be given a behavioral interpretation in terms of gambling~\citep[][Section 2.9.4]{Walley1991}.

A closely related dissimilarity function is $d^*_{\mathrm{COR}}$, defined for all $P, Q \in \mathbb{P}(\mathcal{Y},\mathcal{X})$ by
\begin{align}
    d^*_{\mathrm{COR}}(P, Q) :&= 
    \sup_{\substack{A,B\in\mathcal{A}\\Q(A)>0,P(B)>0}}\frac{P(A)Q(B)}{Q(A)P(B)}\notag \\
    &=1-\frac{1}{d_{\mathrm{COR}}(P, Q)}.
\end{align}
In terms of the densities $p$ and $q$, this simplifies to
\begin{align}
d^*_{\mathrm{COR}}(P, Q) :&=\esssup\frac{p}{q}\esssup\frac{q}{p} \nonumber \\
&=\frac{\esssup\frac{q}{p}}{\essinf\frac{q}{p}}
=\frac{\esssup L}{\essinf L},\label{eq:d_COR_simplified}
\end{align}
with 
\begin{align*}
\esssup f := \inf\{a \in \mathbb{R}: &f \le a \enspace \mu\text{-a.s.}\},\\
\essinf f := \sup\{a \in \mathbb{R}: &f \ge a \enspace \mu\text{-a.s.}\}
\end{align*}
$L=\frac{dQ}{dP}=\frac{q}{p}$ the likelihood ratio of $Q$ w.r.t. $P$ \citep{dumbgen2021}. The $\esssup$ (respectively $\essinf$) is a version of the supremum (infimum) for functions that are only uniquely defined up to measure zero.

Since the distance function  $d_{\mathrm{COR}}$ and dissimilarity function $d^*_{\mathrm{COR}}$ are monotone transformations of each other, the same is true for the resulting robustness metrics: for all $x\in\mathcal{X}$,
\begin{align}
      r_{d_{\mathrm{COR}}}(x) &= 1 - \frac{1}{r_{d^*_\mathrm{COR}}(x)}\label{eq:dcorfromdstarcor}\\
      \intertext{ and }
      r_{d^*_\mathrm{COR}}(x)&=\frac{1}{1-r_{d_\mathrm{COR}}(x)}.
\end{align}

For this reason, we can equivalently work with either of these dissimilarity functions and their corresponding robustness metrics. We will work with $d^*_{\mathrm{COR}}$ and $r_{d^*_{\mathrm{COR}}}$ in our proofs, out of convenience, but will mostly focus on $d_{\mathrm{COR}}$ and $r_{d_\mathrm{COR}}$ in the remainder of the paper because we find $r_{d_\mathrm{COR}}$ more intuitive to interpret.

The following result, the proof of which is available in \autoref{sec:proof}, provides closed-form expressions for the robustness metrics $r_{d^*_{\mathrm{COR}}}$ and $r_{d_{\mathrm{COR}}}$.

\begin{theorem}\label{thm:rob}
Consider a measure $P$ in $\mathcal{P}(\mathcal{Y},\mathcal{X})$, its corresponding density $p$ and, for any given features $x\in\mathcal{X}$, the most and second most likely class according to $p$ given $x$:
\begin{equation*}
    \hat{y}_1\in\arg\max_{y\in\mathcal{Y}}p(y\vert x)
    \text{~~and~~}
    \hat{y}_2\in\arg\max_{y\in\mathcal{Y}\setminus\{\hat{y}_1\}}p(y\vert x).
\end{equation*}
Then the robustness of the prediction $\hat{y}_1$ w.r.t. $d^*_{\mathrm{COR}}$ and $d_{\mathrm{COR}}$ is given by, respectively,
\begin{align}
      r_{d^*_{\mathrm{COR}}}(x) &=\frac{p(\hat{y}_1,x)}{p(\hat{y}_2,x)}\label{eq:rob_ratio}\\
      \intertext{and}
      r_{d_{\mathrm{COR}}}(x)&=\frac{p(\hat{y}_1,x)-p(\hat{y}_2,x)}{p(\hat{y}_1,x)}. \label{eq:rob_prop}
\end{align}
\end{theorem}

This result has a number of noteworthy consequences that are presented below. To the best of our knowledge, no other available robustness metric possesses all of these desirable properties simultaneously, further justifying their adoption.

\paragraph{Easy to compute.}Since both metrics have simple closed-form expressions, this makes them particularly easy to evaluate, whereas existing local robustness metrics typically require performing optimization procedures \citep{correia2019, correia2020rob}. 

\paragraph{Applicable to any type of features.} Unlike global robustness metrics based on epsilon-contamination, our metrics are meaningful (not automatically zero) regardless of whether the features are discrete, continuous or mixed.

\paragraph{Compatible with discriminative models.} A simple application of Bayes' rule allows us to rewrite the obtained expressions in terms of conditional probabilities:
\begin{align}
      r_{d^*_{\mathrm{COR}}}(x) &=\frac{p(\hat{y}_1\vert x)}{p(\hat{y}_2\vert x)}\label{eq:rob_ratio_conditional}\\
      \intertext{and}
      r_{d_{\mathrm{COR}}}(x)&=\frac{p(\hat{y}_1\vert x)-p(\hat{y}_2\vert x)}{p(\hat{y}_1\vert x)}. \label{eq:rob_prop_conditional}
\end{align}
This implies that these metrics can be applied to any machine learning architecture that leads to a discriminative model, that is, one for which only the conditional distribution $p(\cdot|x)$ is available. Our robustness metrics are the first in the literature endowed with such capability.

\paragraph{Easy to interpret.} Since both expressions are based solely on the (conditional or joint) probability of the first and second most probable class, they are extremely simple to interpret, even without detailed knowledge about the reasoning that led to them. In particular, we personally very much like the interpretation of $r_{d_{\mathrm{COR}}}$ as the probability difference between the first and second most probable class, relative to the absolute probability of the most probable one.

\begin{figure}[!tb]
\begin{subfigure}{.49\textwidth}
  \centering
  \includegraphics[width=.95\linewidth]{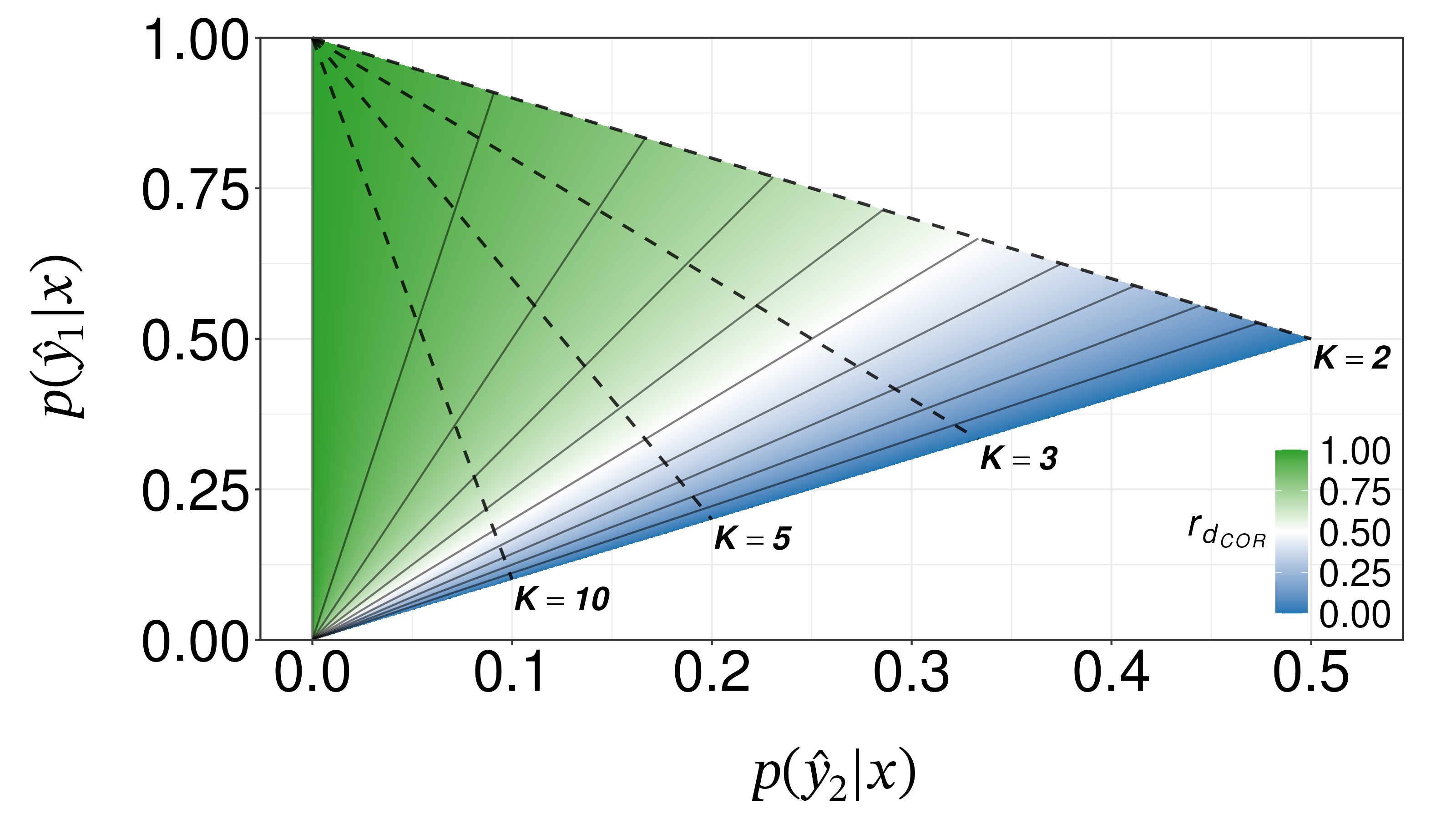}
  \caption{Probability of the second best class}\label{sfig:con_prob}
\end{subfigure}
\begin{subfigure}{.49\textwidth}
  \centering
  \includegraphics[width=.95\linewidth]{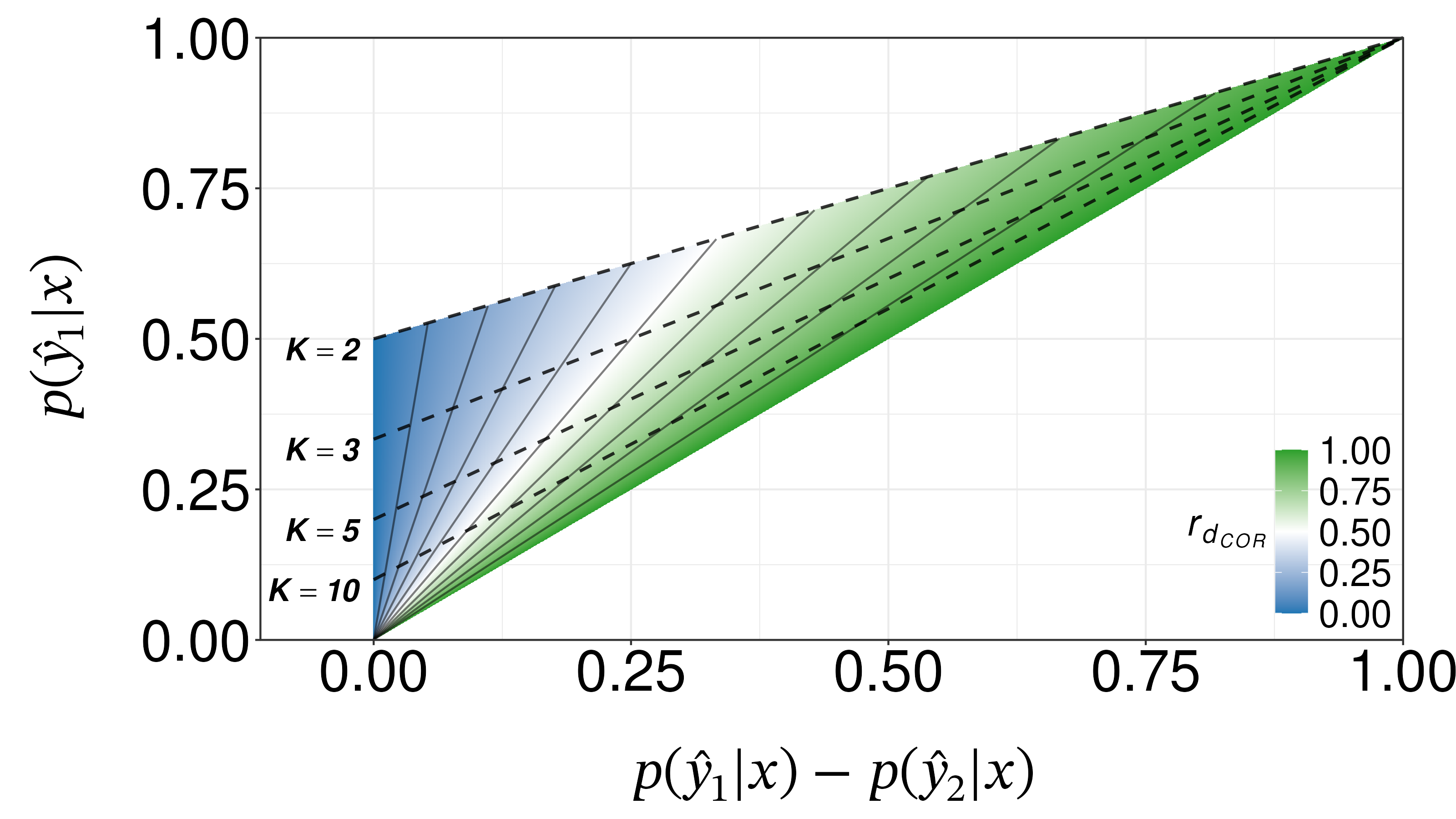}
  \caption{Difference of the top 2 probabilities}\label{sfig:con_diff}
\end{subfigure}
\caption{Contour plots of $r_{d_{\mathrm{COR}}}$ as a function of to $p(\hat{y}_1|x)$ and some other variable. The dashed lines are the lower bounds for when $Y$ has $K$ categories.}
\label{fig:contour}
\end{figure}

To provide some insight into the interpretation of our metric, \autoref{fig:contour} presents two contour plots of $r_{d_{\mathrm{COR}}}(x)$ based on the values of the top probability and either $p(\hat{y}_2 | x)$ or $p(\hat{y}_1 | x) - p(\hat{y}_2 | x)$. The values one can choose for $p(\hat{y}_1 | x)$ and the other entry are restricted by the number of classes in $\mathcal{Y}$, which we denote by $K$. For a fixed value of $K$, the figures present a dashed line associated to it, and any point in the colored region above this line obeys the restrictions imposed by the setting.

In \autoref{sfig:con_prob}, we make a direct study of how the top two probabilities affect robustness. As expected, robustness is high when $p(\hat{y}_1 | x)$ and and $p(\hat{y}_2 | x)$ are far apart, but how distant from each other they have to be depends on how low $p(\hat{y}_2 | x)$ is: it is not the absolute difference that matters, but by how big a factor they differ. So even if  $p(\hat{y}_1 | x)$ is small, robustness can still be high provided $p(\hat{y}_2 | x)$ is only a small fraction of $p(\hat{y}_1 | x)$. This is only possible if $K$ is sufficiently high though because $p(\hat{y}_2 | x)$ being low implies that all other classes also have low probabilities associated to them, while the sum of all probabilities still needs to be one.

Another point of view is presented in \autoref{sfig:con_diff}, where we switch $p(\hat{y}_2 | x)$ for $p(\hat{y}_1 | x) - p(\hat{y}_2 | x)$, making the figure more directly related to Equation~\eqref{eq:rob_prop_conditional} and to two metrics commonly used to evaluate the reliability of predictions: the maximum probability (y-axis) and the confidence gap (x-axis). While smaller values of $p(\hat{y}_1 | x) - p(\hat{y}_2 | x)$ lead to low robustness when $K$ is small, this ceases to be the case for higher $K$ as long as $p(\hat{y}_1 | x)$ is close to the confidence gap (for points close to the diagonal). This demonstrates that our robustness metric can behave in fundamentally different ways depending on the number of classes in $\mathcal{Y}$, and that evaluating only $p(\hat{y}_1 | x)$ or $p(\hat{y}_1 | x) - p(\hat{y}_2 | x)$ is insufficient for capturing this feature.

We end this section by adding a caveat to our robustness metric. Similarly to many of the metrics used in uncertainty quantification (see section 4 of \cite{detavernier2025rob} for some examples), $r_{d_{\mathrm{COR}}}$ may favor overconfident predictions, where the model assigns high probability to a single class even though that is not reflected in the data. 
We therefore recommend to first apply procedures that enhance the overall reliability of the model whenever possible---such as calibration and evaluating overall performance---and to determine the robustness of the individual predictions of such a model in a secondary step.

\section{Experiments}\label{sec:experiments}

The experiments undertaken in this work have two objectives: \textbf{(i)} demonstrate that our reliability metric(s) are capable of assessing the reliability of the predictions of classifiers (section \ref{sec:corr}) and \textbf{(ii)} offer one possible application in which we use the notion of robustness to develop new Dynamic Classifier Selection methods (section \ref{sec:ds}). Detailed information about all datasets in our experiments is given in \autoref{tab:data_info}; the sources were the OpenML \citep{openml}, UCI \citep{uci} and PMLB \citep{romano2021} benchmark repositories. Code reproducing the results can be found in \url{https://github.com/rflassance/rob4discriminative}.

\begin{table}[!tb]
    \centering
    \caption{Details about the datasets in our experiments.}\label{tab:data_info}
    \scriptsize
    \begin{tabular}{ll|rcl}
        Label & Name & $n$ & \#features (\%cont.) & Source\\
        \hline
        $D_1$   & authent               & 1372    & 4 (100\%) &\href{https://www.openml.org/search?type=data&status=active&id=1462}{OpenML}\\
        $D_2$   & bank-additional-full  & 41188   & 20 (25\%) &\href{https://www.openml.org/search?type=data&status=any&id=42813}{OpenML}\\
        $D_3$   & diabetes              & 768     & 8 (25\%) &\href{https://www.openml.org/search?type=data&status=any&id=42608}{OpenML}\\
        $D_4$   & electricity           & 45312   & 8 (88\%) &\href{https://www.openml.org/search?type=data&status=any&sort=qualities.NumberOfInstances&id=151}{OpenML}\\
        $D_5$   & gesture               & 9873    & 32 (100\%) &\href{https://www.openml.org/search?type=data&status=any&sort=match&id=4538}{OpenML}\\
        $D_6$   & magic                 & 19020   & 10 (100\%) &\href{https://www.openml.org/search?type=data&status=any&sort=match&id=40679}{OpenML}\\
        $D_7$   & robot                 & 5456    & 24 (100\%) &\href{https://www.openml.org/search?type=data&status=any&sort=qualities.NumberOfFeatures&id=1497}{OpenML}\\
        $D_8$   & segment               & 2310    & 19 (84\%) &\href{https://www.openml.org/search?type=data&status=any&sort=qualities.NumberOfFeatures&id=958}{OpenML}\\
        $D_9$   & students              & 4424    & 36 (19\%) & \href{https://archive.ics.uci.edu/dataset/697/predict+students+dropout+and+academic+success}{UCI}\\
        $D_{10}$& texture               & 5500    & 40 (100\%) &\href{https://www.openml.org/search?type=data&status=any&sort=qualities.NumberOfFeatures&id=40499}{OpenML}\\
        $D_{11}$& twonorm               & 7400    & 20 (100\%) &\href{https://www.openml.org/search?type=data&status=any&sort=qualities.NumberOfFeatures&id=1507}{OpenML}\\
        $D_{12}$& vowel                 & 990     & 12 (83\%) &\href{https://www.openml.org/search?type=data&status=any&sort=qualities.referencedNumberOfFeatures&id=1016}{OpenML}\\
        $D_{13}$& waveform\_21          & 5000    & 21 (100\%) & \href{https://epistasislab.github.io/pmlb/profile/waveform_21.html}{PMLB} \\
        $D_{14}$& waveform\_40          & 5000    & 40 (100\%) & \href{https://epistasislab.github.io/pmlb/profile/waveform_40.html}{PMLB} \\
        $D_{15}$& wdbc                  & 569     & 30 (100\%) &\href{https://www.openml.org/search?type=data&status=any&sort=qualities.NumberOfFeatures&id=1510}{OpenML}\\
        \hline
    \end{tabular}
\end{table}

\subsection{Correlation with Accuracy}\label{sec:corr}

To assess the quality of our robustness metric(s), we follow \citet{detavernier2025rob, detavernier2025comb} in using Accuracy Rejection Curves \citep[ARC,][]{nadeem2009}. To generate such an ARC, we first order the data according to a specific strategy, which in this case is in order of increasing values of robustness. Next, we evaluate the accuracy of the model in regards to the ordered data, gradually throwing away the first samples (with low robustness, in this case) and recalculating the accuracy for the remaining data, leading to an accuracy curve indexed by the proportion of samples that were removed (the rejection rate). Examples of such ARCs, which we'll analyse in the next section, can be seen in \autoref{fig:robs}. When the ordering criteria is good, the accuracy of the model quickly goes to 1 as we disregard samples. Conversely, when the ordering is not good, the ARC remains closer to the starting accuracy. This way, ARCs provide a visual method for  evaluating how well a robustness metric is at correlating with accuracy. In sections \ref{sec:corr_robs} and \ref{sec:corr_mods}, we perform 15 train-test splits for each dataset analyzed. Then, for every split, we obtain the ARC for the test set and present the average of the ARCs as our result.

\subsubsection{Comparison Between Robustness Metrics}\label{sec:corr_robs}

Our first experiment compares how our robustness metric, $r_{d_{COR}}$, fares when compared to an alternative. Considering that part of the focus of this work is to provide a robustness metric that remains meaningful when the datasets possess continuous features, we turn our attention to another method with a similar capability: a local robustness metric based on Generative Forests \citep[GeF,][]{correia2020rob} that we refer to as $r_{GeF}$. This metric is specifically designed for GeFs--- which are a generative version of Random Forests \citep{breiman2001}---and epsilon-contaminates the parameters of a Generative Forest to assess robustness.

\begin{figure}[!tb]
\begin{subfigure}{.49\textwidth}
  \centering
  \includegraphics[width=.9\linewidth]{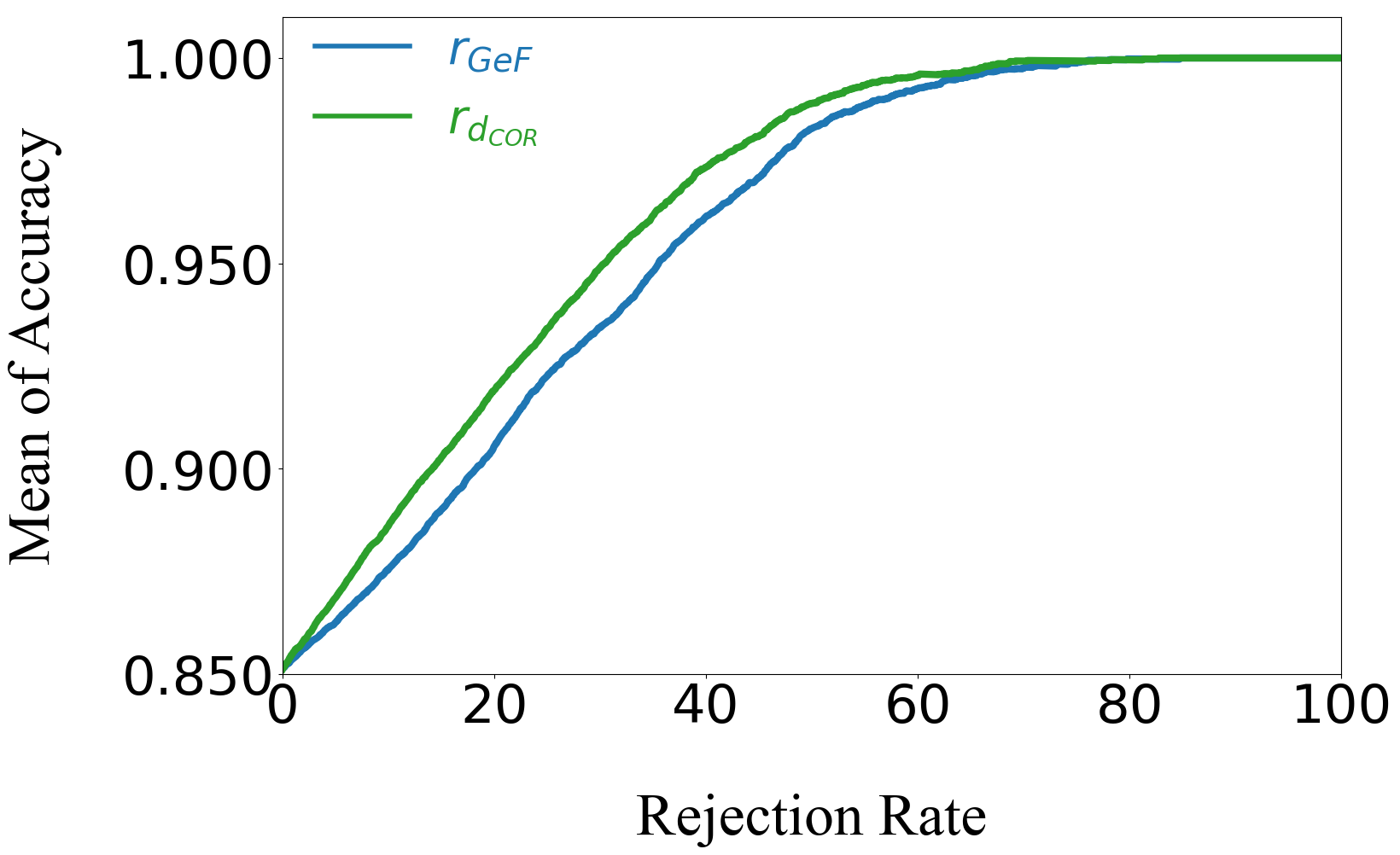}
  \caption{waveform\_21 dataset ($D_{13}$)}\label{sfig:wave}
\end{subfigure}
\begin{subfigure}{.49\textwidth}
  \centering
  \includegraphics[width=.9\linewidth]{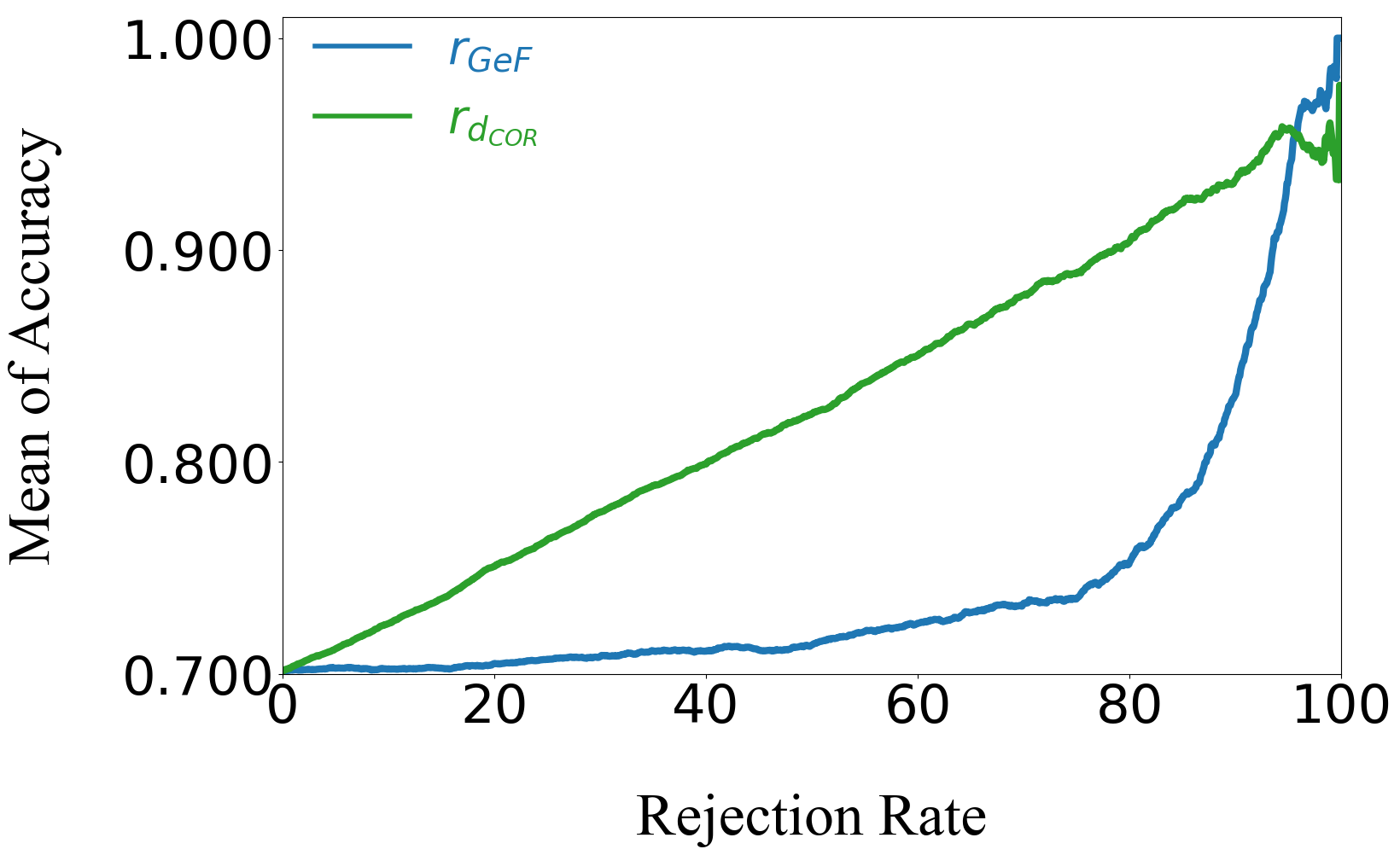}
  \caption{students dataset ($D_9$)}\label{sfig:student}
\end{subfigure}
\caption{Comparison between the ARCs of $r_{GeF}$ and $r_{d_{COR}}$ for two datasets.}
\label{fig:robs}
\end{figure}

\autoref{fig:robs} shows the ARC curves of both robustness metrics for two different datasets, as an illustration of two extreme situations that occur throughout the datasets. In \autoref{sfig:wave}, we note that both robustness metrics provide a good ordering criteria for the data, with $r_{d_{COR}}$ slightly outperforming $r_{GeF}$. In the other extreme, \autoref{sfig:student} highlights a situation in which $r_{GeF}$ is doing a poor job at correlating with accuracy, whereas $r_{d_{COR}}$ remains more consistent. Considering that in both cases $r_{d_{COR}}$ has a better performance, and that we observed similar behaviour in the other datasets, this leads us to disregard the use of $r_{GeF}$ in further analyses and focus solely on $r_{d_{COR}}$.

\subsubsection{Comparison Between Models}\label{sec:corr_mods}

Unlike $r_{GeF}$, our metric $r_{d_{COR}}$ is not restricted to Generative Forests, being applicable to any generative or discriminative model of interest. Hence, we can also make use of ARCs to verify if our robustness metric correlates with accuracy for different types of models. In our next experiment, we do this for Gradient Boosting (GB), Random Forests (RF), XGBoost \citep[XGB,][]{chen2016} and GeFs.

\begin{figure}[!tb]
\begin{subfigure}{.49\textwidth}
  \centering
  \includegraphics[width=.9\linewidth]{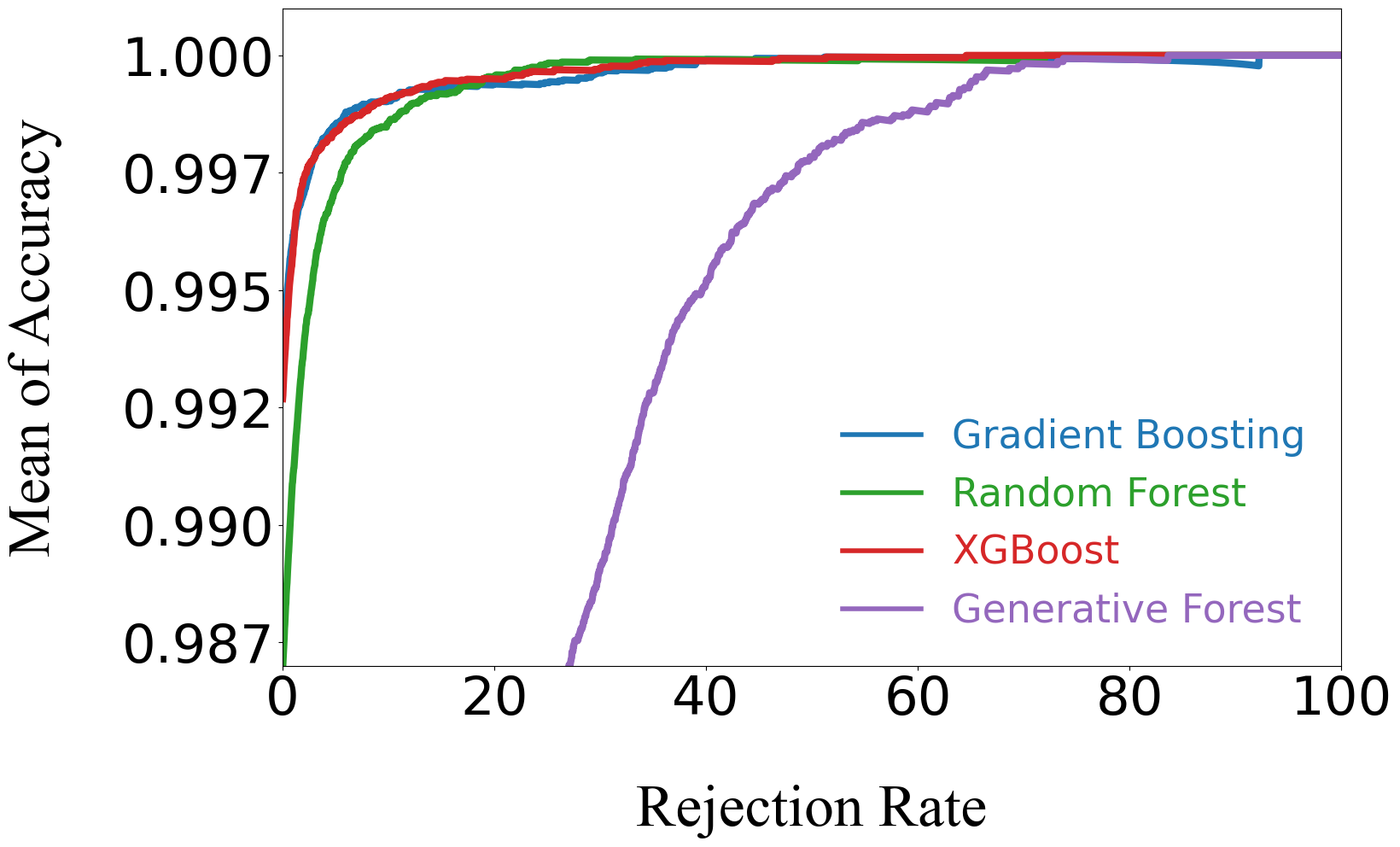}
  \caption{robot dataset ($D_{7}$)}\label{sfig:robot_rob}
\end{subfigure}
\begin{subfigure}{.49\textwidth}
  \centering
  \includegraphics[width=.9\linewidth]{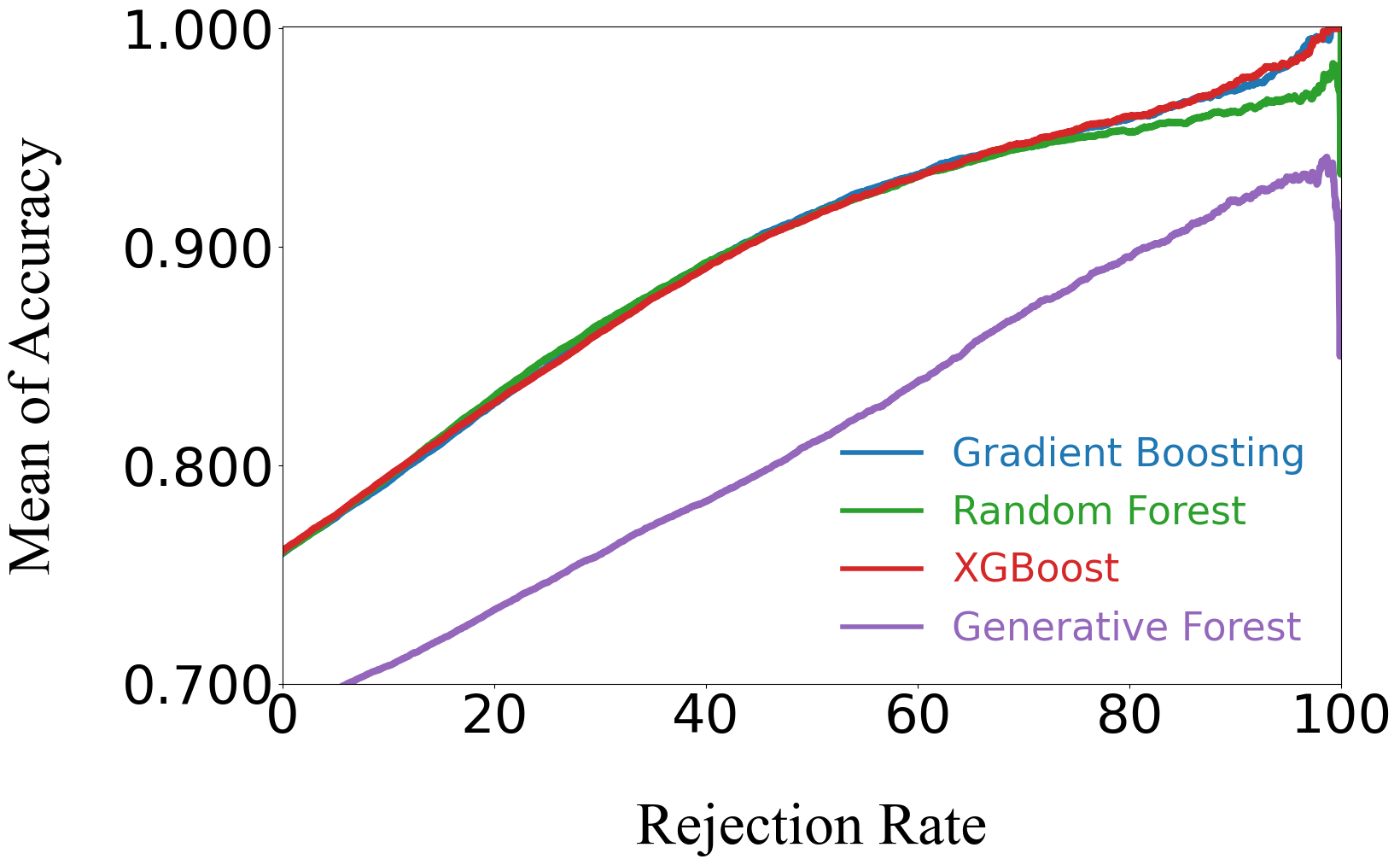}
  \caption{students dataset ($D_9$)}\label{sfig:stu_rob}
\end{subfigure}
\caption{Comparison between the ARCs of different models for two datasets when ordered by $r_{d_{COR}}$.}
\label{fig:mod_robs}
\end{figure}

\autoref{fig:mod_robs} presents the ARC of these four models in specific datasets. The leftmost value of the ARCs is the base accuracy of the models, indicating that Generative Forests was the least competitive model in both settings. For every model and both datasets, we observe that our robustness metric $r_{d_{COR}}$ still correlates with accuracy, but that the growth of the ARC varies based on dataset and model. In \autoref{sfig:robot_rob}, the ARCs of the GB and of the XGB are similar at the start, being both superseded by the RF around a rejection rate of 20\% (even though the RF started out with a lower base accuracy) and then exhibiting about the same behavior as RF around 40\%, while the GeF only catches up around 70\%. As for \autoref{sfig:stu_rob}, the top three models initially perform equally well, with the GB and XGB eventually outperforming the other models around a rejection rate of 70\%, while the GeF is again not competitive. These results not only imply that robustness correlates with accuracy, but also that different models might be the top performers at specific rejection rates. Moreover, the GeF underperformed in both cases, leading us to disregard it from further analyses.

One thing for which results like these can be used, is to gain an understanding of how well robustness is capable of assessing the reliability of the predictions of different models. Another, however, is to use them to increase performance in terms of accuracy. Most obviously, in a context where we can allow ourselves to reject a percentage of the data---for example because we can label those manually---we can use robustness-based ARCs to determine wich percentage to reject, and to determine which model performs best on the remaining data. However, it is also possible to use robustness to increase performance in contexts where rejection is not an option. This is what we come to next.

\subsection{Dynamic Selection of Classifiers}\label{sec:ds}

To demonstrate the potential use of robustness metrics, we propose two robustness-based strategies (RS-D and RS-I) for Dynamic Selection (DS) of classifiers and compare them to other strategies in benchmark datasets. DS is an umbrella term for techniques that combine or choose between multiple base models, contingent on the set of features being used as input \citep{cruz2018}. Hence, depending on the values of the new sample, a DS classifier uses different base models as reference to provide a prediction.

\begin{figure}[!tb]
\begin{subfigure}{.24\textwidth}
  \centering
  \includegraphics[width=.9\linewidth]{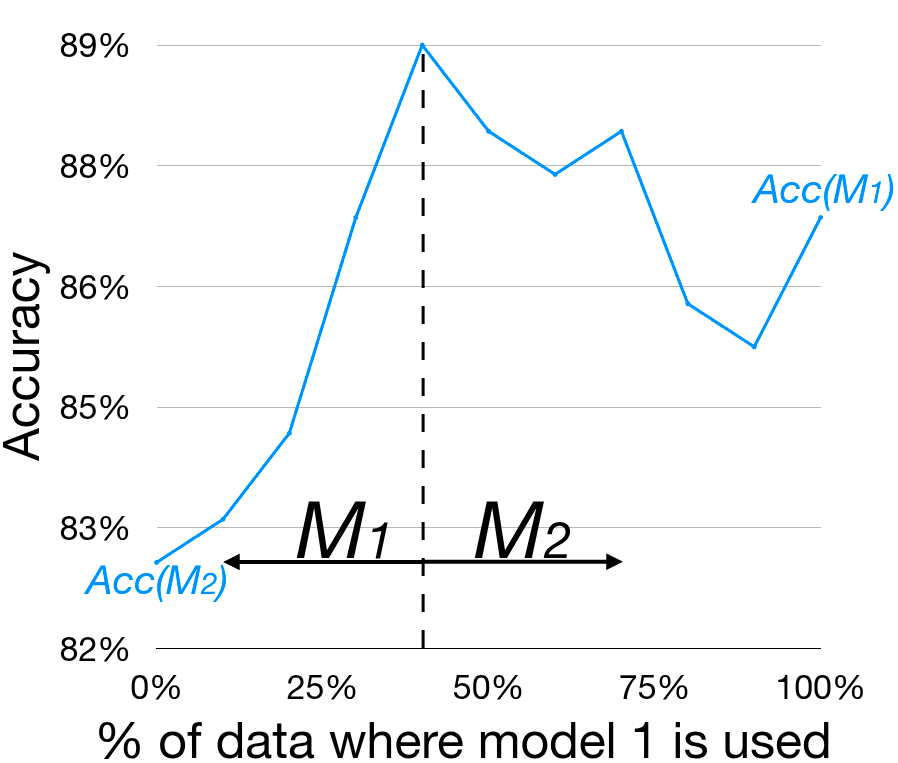}
  \caption{RS-D}\label{sfig:R1}
\end{subfigure}%
\begin{subfigure}{.24\textwidth}
  \centering
  \includegraphics[width=.9\linewidth]{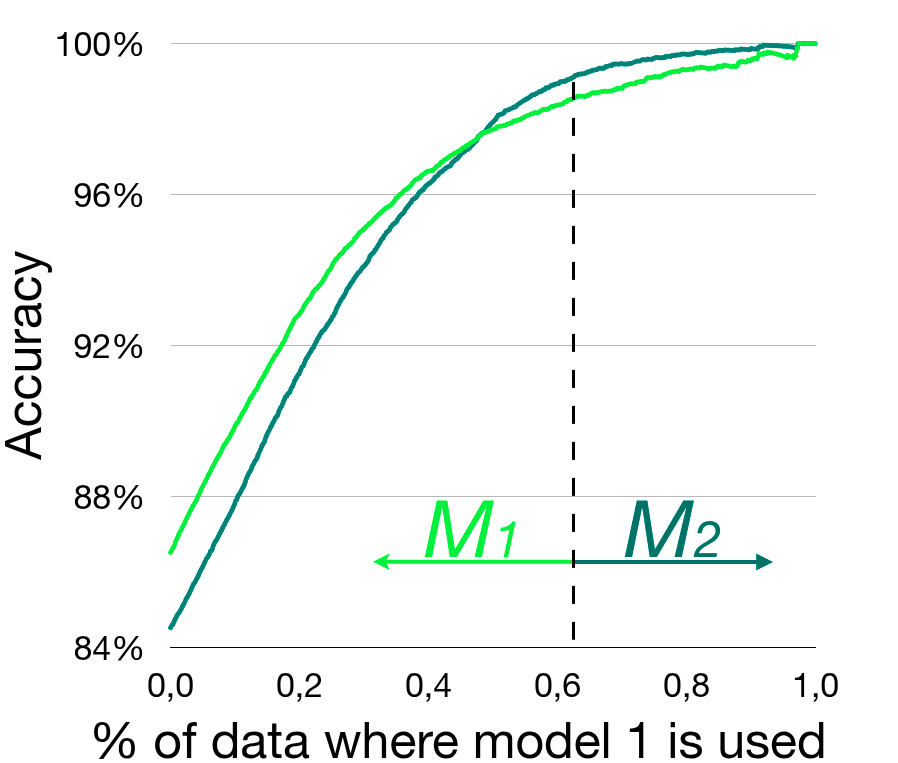}
  \caption{RS-I}\label{sfig:R2}
\end{subfigure}
\caption{Robustness-based strategies for DS. $M_1$ is the model with the best accuracy for the validation set, while $M_2$ is the second best.}
\label{fig:strats}
\end{figure}

\begin{table*}
    \centering
    \caption{Mean accuracy of each method for different datasets.\\
    \small  Best accuracy in bold, second best underlined.}\label{tab:mean_acc}
    \fontsize{8.7pt}{9pt}\selectfont
    \begin{tabular}{l|ccc|ccccccc}
        Label &GB&RF&XGB &SB & MCB & KNORA-U & KNORA-E & META-DES & RS-D & RS-I\\
        \hline
        $D_{1}$ &0.99742 & 0.99356 & 0.99388&\textbf{0.9971} & \underline{0.99678} & 0.99614 & \underline{0.99678} & 0.99581 & 0.99549 & \underline{0.99678} \\
        $D_{2}$ &0.91598 & 0.91489 & 0.91594&0.91569 & 0.91568 & 0.9158  & 0.91573 & \textbf{0.91657} & \underline{0.91594} & 0.91567\\
        $D_{3}$ &0.77126 & 0.76954 & 0.75517&0.76264 & 0.76494 & \underline{0.76667} & \underline{0.76667} & \underline{0.76667} & \textbf{0.76724} & 0.76149\\
        $D_{4}$ &0.86772 & 0.91021 & 0.88572&\underline{0.91021} & 0.89522 & 0.89445 & 0.89647 & 0.90386 & 0.90975 & \textbf{0.91033}\\
        $D_{5}$ &0.60909 & 0.66284 & 0.65367&0.66284 & 0.6525  & 0.6516  & 0.65744 & 0.65268 & \textbf{0.6664}  & \underline{0.66451}\\
        $D_{6}$ &0.88019 & 0.88215 & 0.88036&0.88089 & 0.88024 & 0.88199 & 0.8819  & \textbf{0.88288} & \underline{0.88227} & 0.88106\\
        $D_{7}$ &0.9974  & 0.99333 & 0.9965 &0.99683 & 0.9956  & 0.99691 & 0.99683 & \underline{0.99707} & \textbf{0.9978}  & 0.99699\\
        $D_{8}$ &0.97406 & 0.97387 & 0.97483&0.97426 & 0.97426 & 0.97483 & 0.97483 & \underline{0.97522} & \textbf{0.97618} & 0.97483\\
        $D_{9}$ &0.76581 & 0.76752 & 0.77023&0.77013 & 0.76842 & 0.77033 & 0.77033 & \textbf{0.77193} & \underline{0.77073} & \underline{0.77073}\\
        $D_{10}$&0.98079 & 0.9753  & 0.98103&0.98047 & 0.97942 & 0.982   & 0.98184 & \underline{0.98208} & \textbf{0.98257} & 0.98184\\
        $D_{11}$&0.97036 & 0.9721  & 0.97006& 0.9709  & 0.97126 & \textbf{0.9733}  & 0.97306 & \textbf{0.9733}  & 0.97234 & 0.97174\\
        $D_{12}$&0.86846 & 0.95615 & 0.90022&\textbf{0.95615} & 0.91767 & 0.92662 & 0.92796 & 0.92617 & 0.94183 & \underline{0.95168}\\
        $D_{13}$&0.85326 & 0.85175 & 0.85885&0.85442 & 0.85477 & \textbf{0.85823} & \underline{0.85761} & \underline{0.85761} & 0.85539 & 0.8561\\
        $D_{14}$&0.85513 & 0.8553  & 0.86143&0.85841 & 0.85628 & \underline{0.86019} & \textbf{0.86045} & 0.85957 & 0.8577  & 0.85965\\
        $D_{15}$&0.94419 & 0.94651 & 0.94729& 0.94496 & \underline{0.94574} & 0.94496 & 0.94186 & 0.94186 & 0.94109 & \textbf{0.94651}\\
        \hline
        Beats SB &5/15&4/15&7/15& - & 4/15 & 10/15 & 9/15 & 10/15 & 10/15 & 11/15\\
        \hline
    \end{tabular}
\end{table*}

Let $M_1$ and $M_2$ be the models with the highest and second highest accuracy in a hold-out validation set and let $r_1(\cdot)$ and $r_2(\cdot)$ be the their robustness metrics (since we focus solely on $r_{d_\mathrm{COR}}$, we drop the subscript $d_\mathrm{COR}$ for notational convenience). For both DS classifier strategies, we start by ordering the validation set based on $\nicefrac{r_2(\cdot)}{r_1(\cdot)}$. For instances that appear towards the beginning of the ordering, the prediction $M_1$ is more robust than that of $M_2$, and vice versa for instances towards the end of the ordering. 
The idea is now to only favor $M_2$ over $M_1$ when its robustness is considerably above that of $M_1$, so for instances towards the end of the ordering. Concretely, we choose a threshold $t$ and, for any new instance $x_{\text{new}}$, we predict the corresponding class $y_{\text{new}}$ using $M_2$ instead of $M_1$ only in cases when $\nicefrac{r_2(x_{\text{new}})}{r_1(x_{\text{new}})} > t$, and otherwise use $M_1$. The only difference between RS-D and RS-I is the way in which the threshold $t$ is determined.

\autoref{fig:strats} provides an illustration on how to derive the threshold for each strategy. On the horizontal axis of both graphs, we depict the percentage of the holdout validation data for which $\nicefrac{r_2(x)}{r_1(x)}$ does not exceed the chosen $t$. The vertical axis depicts the accuracy on the same validation data, either for the DS classifier for this choice of $t$ when applied to all validation data  (\autoref{sfig:R1}) or for the individual models when applied to the data that exceeds $t$ (\autoref{sfig:R2}). RS-D (\autoref{sfig:R1}) takes a Direct approach (hence the D). The idea is simply to choose $t$ such that the resulting strategy maximizes the accuracy in the validation set. This is equivalent to finding the proportion $p$ of the validation set with the lowest robustness difference that yields the highest accuracy when using $M_1$ to predict the class for this part of the data and $M_2$ to predict the rest. RS-I (\autoref{sfig:R2}), on the other hand, takes a more indirect approach. The idea here is to compare the ARCs of $M_1$ and $M_2$ for the ordering based on $\nicefrac{r_2(\cdot)}{r_1(\cdot)}$ (so not based on $r_1(\cdot)$ and $r_2(\cdot)$, respectively), and to verify at what point (if at all) the ARC of $M_2$ has the largest accuracy advantage over $M_1$. So we choose $t$ as the point for which, on the data for which $\nicefrac{r_2(x)}{r_1(x)}$ exceeds $t$, $M_2$ provides the greatest accuracy gain compared to $M_1$.

\autoref{tab:mean_acc} presents the mean accuracy of each DS technique for the datasets in \autoref{tab:data_info}. The base models used in the experiments were GB, RF and XGB. For each dataset, 15 different random train-validation-test splits were performed, with respective proportions of 0.7, 0.15 and 0.15. For each random split, the base models were fitted to the training set, with their hyperparameters being selected through grid search with a 5-fold cross-validation using the scikit-learn package \citep{scikit}. Then, the accuracy for the validation data was used for choosing the single best base model (SB) and for tuning the DS strategies. The competing DS methods (MCB, KNORA-U, KNORA-E and META-DES) can be found in \cite{cruz2018} and are based on using KNN on the feature space to come up with the optimal model, using the validation set to decide the number of neighbors (maximum of 10). All the competing DS techniques were implemented through the DESlib library \citep{cruz2020}.

\begin{figure}[!htb]
\begin{subfigure}{.24\textwidth}
  \centering
  \includegraphics[width=.95\linewidth]{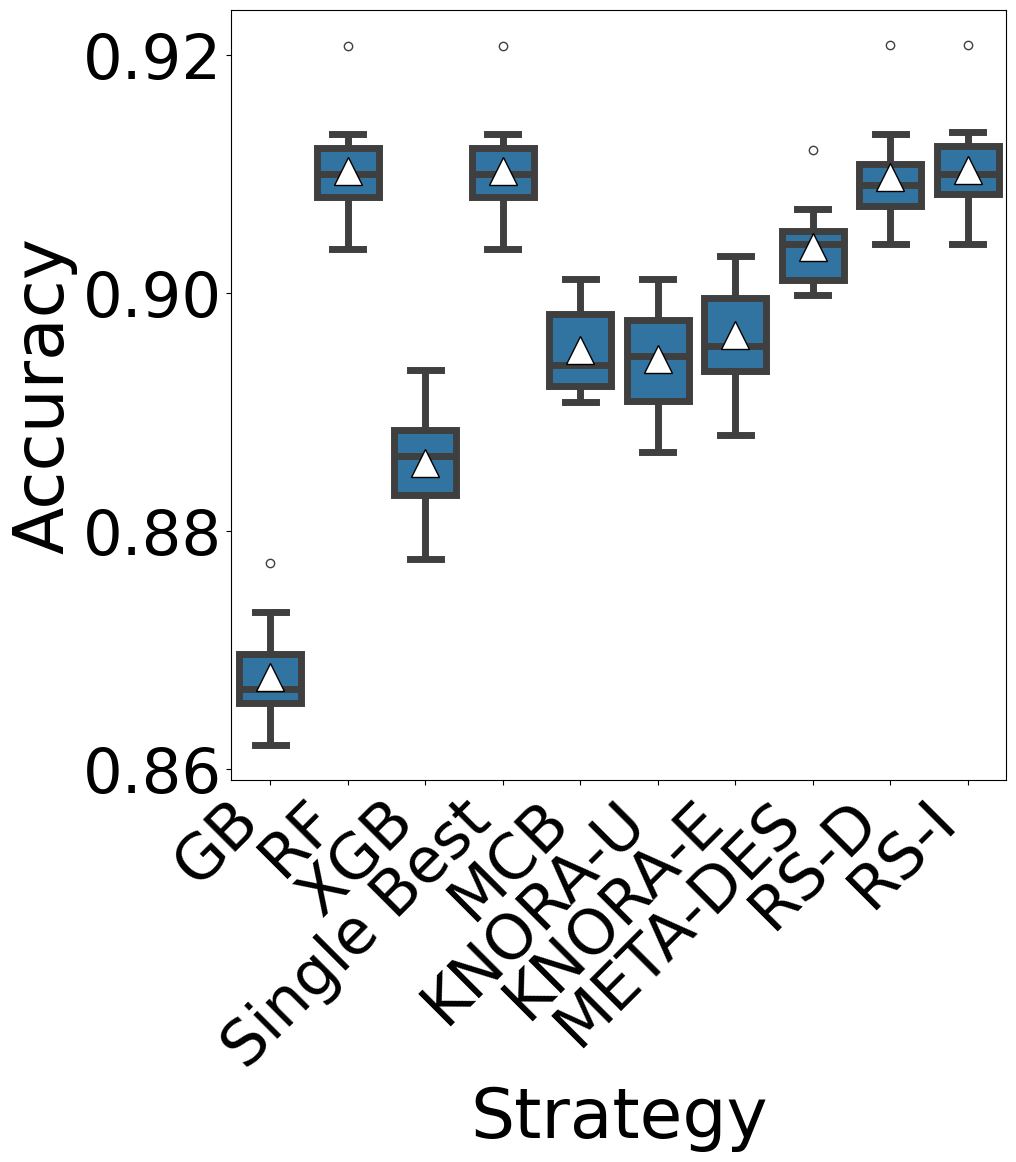}
  \caption{electricity ($D_4$)}
\end{subfigure}%
\begin{subfigure}{.24\textwidth}
  \centering
  \includegraphics[width=.95\linewidth]{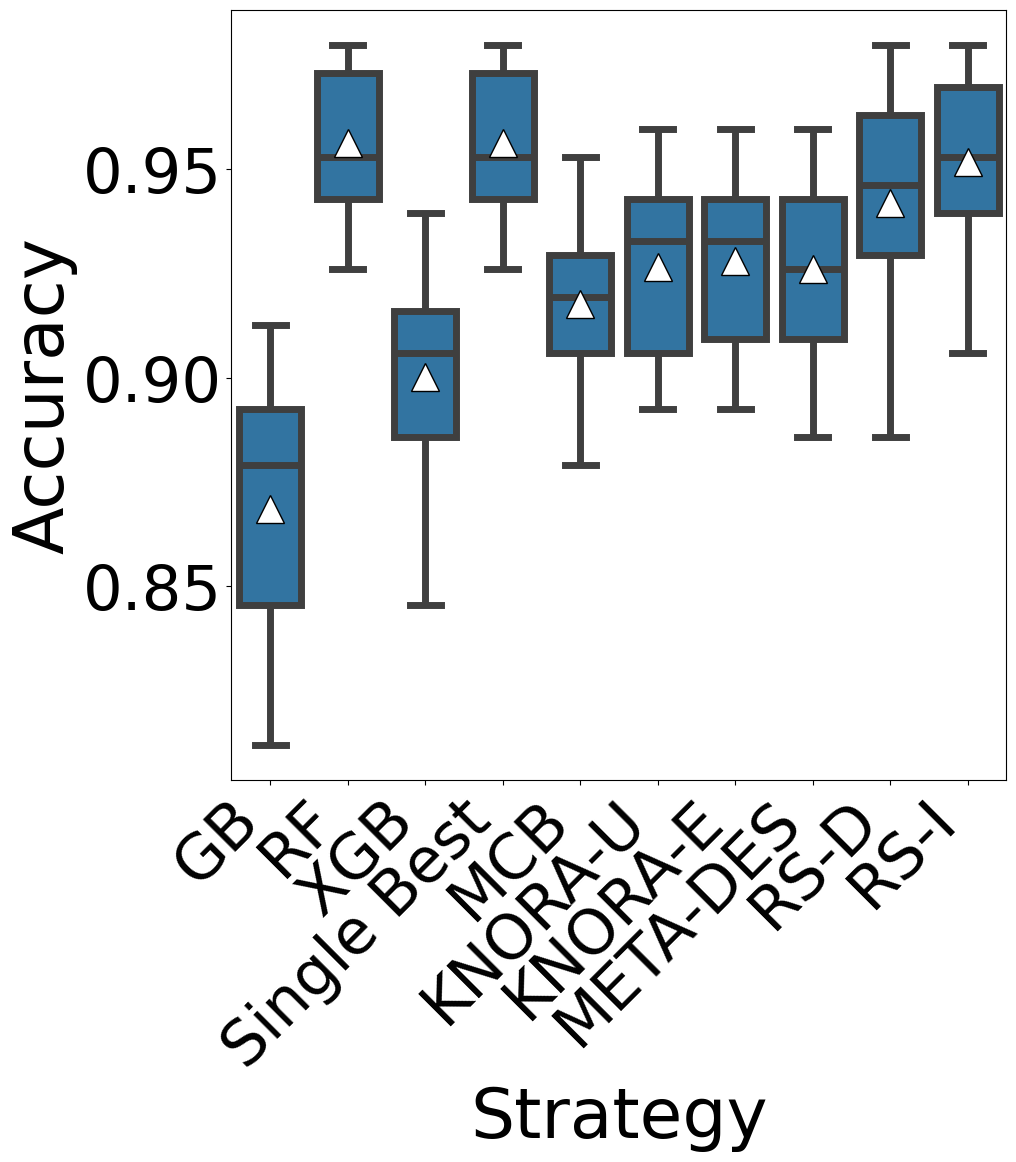}
  \caption{vowel ($D_{12}$)}
\end{subfigure}\\[10pt]
\begin{subfigure}{.24\textwidth}
  \centering
  \includegraphics[width=.95\linewidth]{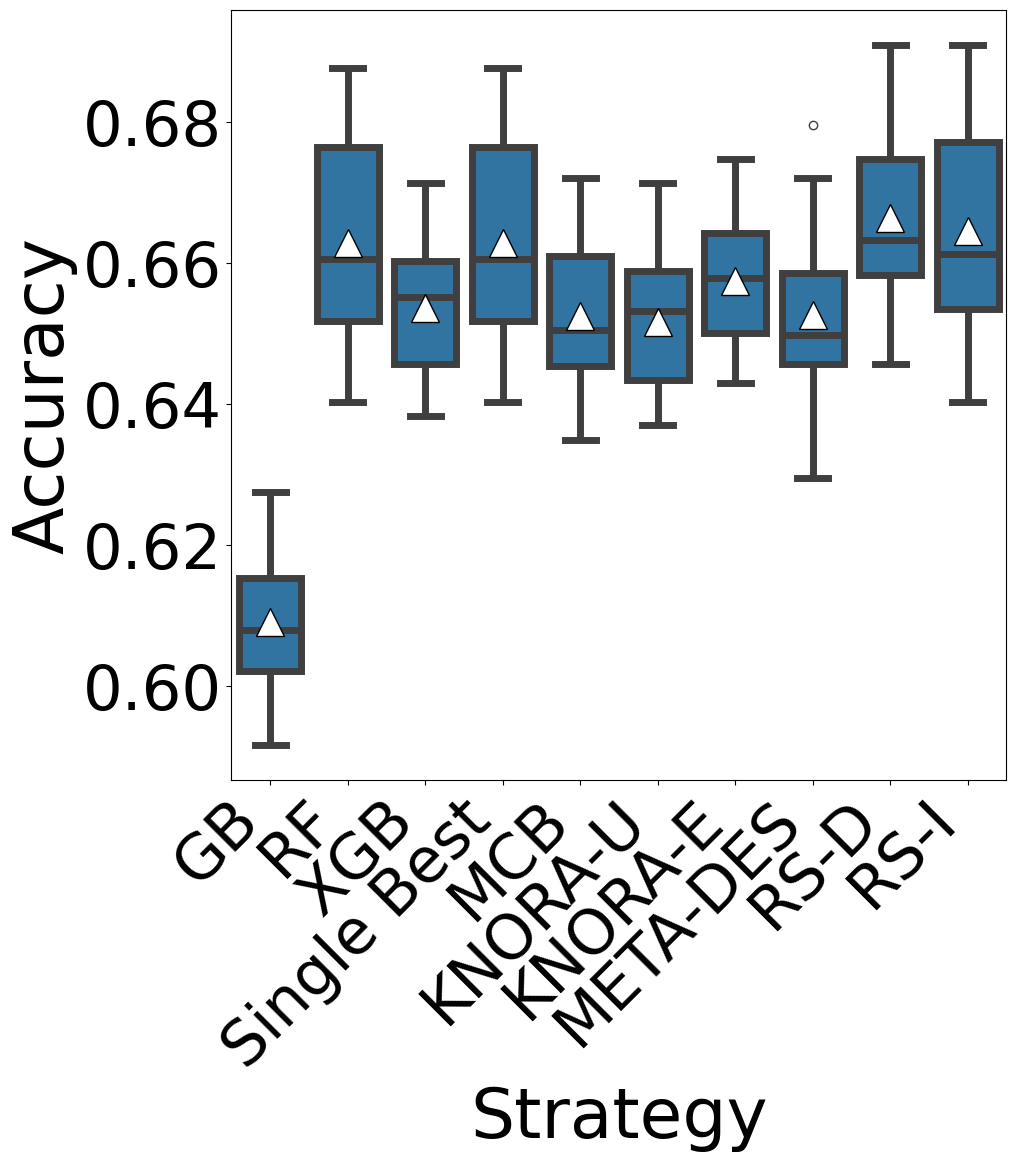}
  \caption{gesture ($D_{5}$)}
\end{subfigure}%
\begin{subfigure}{.24\textwidth}
  \centering
  \includegraphics[width=.95\linewidth]{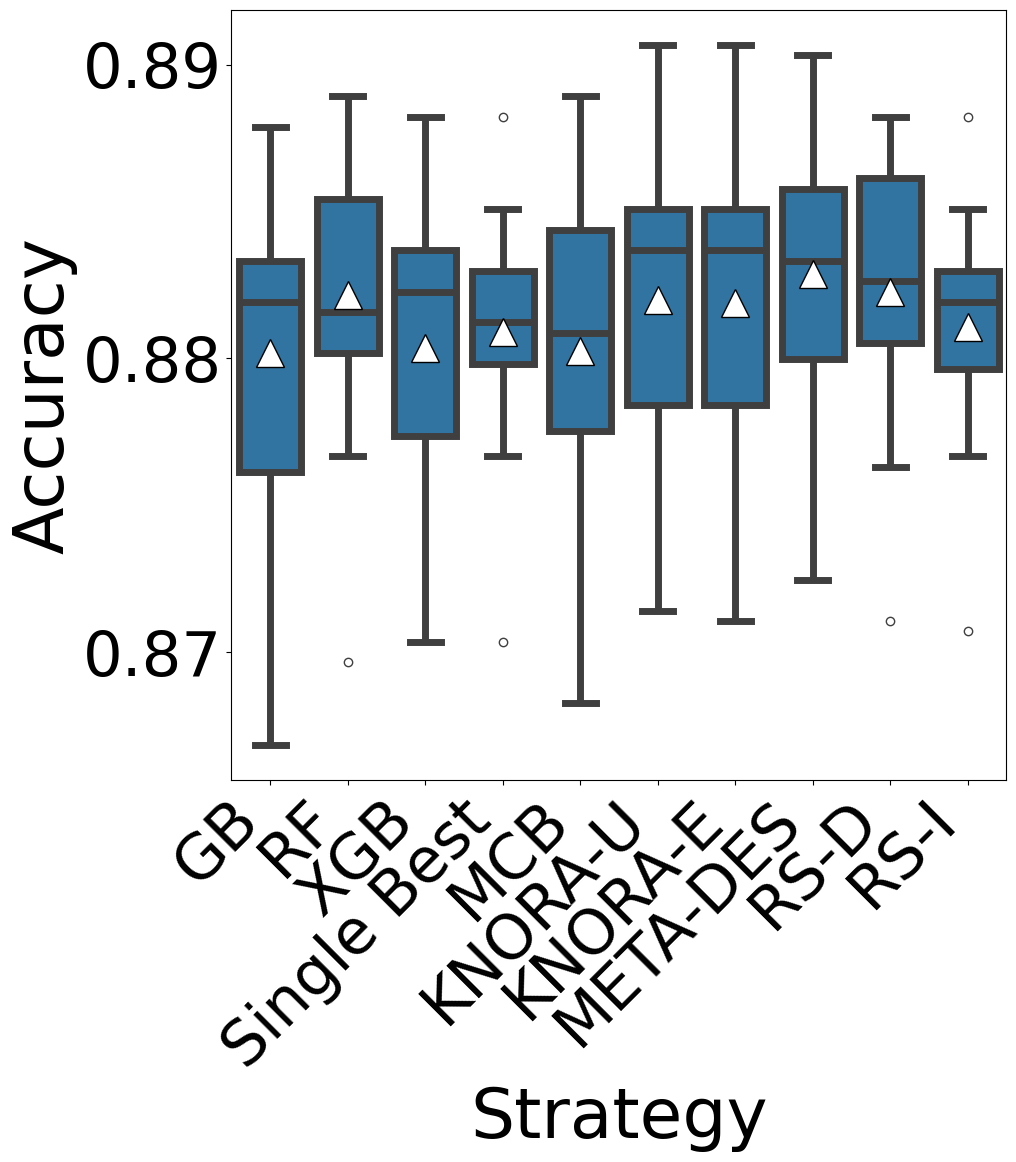}
  \caption{magic ($D_{6}$)}
\end{subfigure}
\caption{Boxplot of accuracies for each strategy and different datasets; white triangles represent the mean.}
\label{fig:boxplots}
\end{figure}

Analysing the results in \autoref{tab:mean_acc}, we see that RS-D is the best performing strategy (indicated in bold) most often. Moreover, RS-D featured as one of the two best strategies in the majority of the datasets, along with META-DES. Furthermore, among the datasets in which RS-D wasn't listed as one of the two best strategies, RS-I was among the best in more than half of them. Compared to the Single Best (SB) strategy, RS-I was the alternative that most often had a better performance against it, even in circumstances where RS-D was not among the two best strategies. Moreover, in cases where one base model clearly outperforms the others---such as in the electricity ($D_4$) and vowel ($D_{12}$) datasets---all competing DS strategies based on the feature space underperform when compared to our robustness-based strategies. \autoref{fig:boxplots} illustrates this situation in the two plots on the first row, while presenting more typical settings in the second row.

\begin{table*}
    \centering
    \caption{Mean accuracy of each method under label corruption ($\rho = 0.05$).\\
    \small  Best accuracy in bold, second best underlined.}\label{tab:mean_acc_noise}
    \fontsize{8.7pt}{9pt}\selectfont
    \begin{tabular}{l|ccc|ccccccc}
        Label &GB&RF&XGB            & SB & MCB & KNORA-U & KNORA-E & META-DES & RS-D & RS-I\\
        \hline
        $D_{1}$  &0.98712 & 0.98583 & 0.99227&\underline{0.99259} & 0.99002 & 0.99163 & 0.9913  & 0.99195 & \underline{0.99259} & \textbf{0.99291} \\
        $D_{2}$  &0.91474 & 0.91421 & 0.91554&0.91501 & 0.9152  & 0.91579 & \underline{0.91607} & \textbf{0.91611} & 0.91549 & 0.91514\\
        $D_{3}$  &0.75747 & 0.76954 & 0.75172&0.76092 & 0.75632 & 0.76264 & \underline{0.76379} & 0.76092 & \textbf{0.76552} & 0.76149\\
        $D_{4}$  &0.86471 & 0.90057 & 0.88439&\underline{0.90057} & 0.88947 & 0.89033 & 0.89175 & 0.89528 & 0.90055 & \textbf{0.90062}\\
        $D_{5}$  &0.59879 & 0.65807 & 0.64948&0.65601 & 0.64444 & 0.64534 & 0.652   & 0.64656 & \textbf{0.66023} & \underline{0.65857}\\
        $D_{6}$  &0.87461 & 0.88106 & 0.87655&0.87783 & 0.87748 & 0.87835 & 0.87767 & \underline{0.87874} & \textbf{0.87993} & 0.87821\\
        $D_{7}$  &0.98926 & 0.96809 & 0.99373&\textbf{0.99373} & 0.98689 & 0.99365 & 0.99365 & \textbf{0.99373} & 0.99292 & \textbf{0.99373}\\
        $D_{8}$  &0.95677 & 0.91873 & 0.97003&\underline{0.96926} & 0.95543 & 0.96619 & 0.96619 & 0.96734 & 0.96849 & \textbf{0.96945}\\
        $D_{9}$  &0.76361 & 0.76942 & 0.76882&0.76632 & 0.76642 & 0.76862 & 0.76872 & \underline{0.76932} & \textbf{0.77043} & 0.76832\\
        $D_{10}$ &0.96578 & 0.87264 & 0.98136&\underline{0.98136} & 0.9632  & 0.977   & 0.97684 & 0.97571 & 0.98055 & \textbf{0.98144}\\
        $D_{11}$ &0.96406 & 0.97138 & 0.96208&0.96952 & 0.9664  & 0.9697  & 0.96946 & \underline{0.96982} & \textbf{0.97024} & \underline{0.96982}\\
        $D_{12}$ &0.85414 & 0.9387  & 0.90291&\textbf{0.93647} & 0.90559 & 0.91902 & 0.91946 & 0.91678 & 0.92975 & \underline{0.93512}\\
        $D_{13}$ &0.8498  & 0.85362 & 0.8522&0.85051 & 0.85024 & \underline{0.85424} & \textbf{0.85619} & 0.85317 & 0.85193 & 0.85078\\
        $D_{14}$ &0.8506  & 0.85353 & 0.85229&0.84953 & 0.85211 & \underline{0.85451} & 0.85335 & 0.85442 & \textbf{0.85486} & 0.8506\\
        $D_{15}$ &0.94031 & 0.94496 & 0.94884&\underline{0.94806} & 0.94186 & 0.94729 & \underline{0.94806} & 0.94729 & \textbf{0.94884} & 0.94729\\
        \hline
        Beats SB&1/15&8/15&6/15& - & 3/15 & 7/15 & 5/15 & 6/15 & 9/15 & 12/15\\
        \hline
    \end{tabular}
\end{table*}

\subsubsection{Performance Under Label Corruption}\label{sec:label_corr}
While \autoref{tab:mean_acc} shows that robustness can be used to build a competitive DS classifier, it only evaluates settings in which the test set comes from the same distribution as the rest of the data. To evaluate the performance of all strategies in a context of distribution shift, we perform label corruption on the training and validation sets, which works by uniformly changing the labels of the response variable for a proportion $\rho$ of the data. This same technique has been previously explored in \cite{li2020}.

In \autoref{tab:mean_acc_noise}, we set $\rho = 0.05$ and repeat the analyses. RS-D and RS-I remain as top performers in the new setting, but this time none of the competing DS strategies (not even META-DES) fares better than choosing the Single Best (SB) base model based on the accuracy in the validation set. Moreover, while RS-D still remains the highest performing strategy, RS-I is the technique that is most often superior to the Single Best. This pattern suggests that RS-I can be seen as the more reliable option, while RS-D is the one with greater chances of offering the best performance. Such behavior remains consistent even when higher levels of label corruption are applied, as demonstrated in Tables \ref{tab:mean_acc_noise10} and \ref{tab:mean_acc_noise20} in the supplementary material.

\section{Discussion}\label{sec:disc}
In this work, we have shown that robustness quantification can be brought to more general settings, being applicable to discriminative models and to settings with continuous features. Still, the choice of the dissimilarity function $d$ of course remains somewhat arbitrary, justifying further studies into what other possible robustness metrics could have such properties.

Section \ref{sec:corr_robs} provides evidence---based on ARCs---that, even though our robustness metric does not take the architecture of the predictive model into account, it still outperforms the competing metric. This situation contrasts the findings of \cite{detavernier2025rob, detavernier2025comb}, which have shown that local robustness metrics can in fact perform better than global metrics. The reason for this discrepancy is still to be determined, with some possibilities being the fact that these works were limited to naive Bayes classifiers or that they were restricted to discrete features.

Considering the results in section \ref{sec:ds}, the use of robustness has presented superior performance in the context of Dynamic Selection of classifiers, especially when dealing with label corruption. One of the possible reasons is that standard DS techniques make assessments on the entirety of the feature space, which can lead to instability, whereas robustness offers a lower dimensional but rich representation of the data, leading to more stable procedures. Conversely, combining evaluations in the robustness space in a similar manner as that of other DS techniques could also offer improvements, being a potential topic for further research.

Besides robustness, there are other reliability metrics that could also offer a lower dimensional representation of the feature space, such as those used in uncertainty quantification \citep{hullermeier2021}, which could therefore be applied to the same Dynamic Selection tasks. More than arguing for the use of one over the other, we believe that each can bring different aspects of the model and the data to the spotlight, and should thus be tried in combination. In fact, combining robustness and uncertainty in a single metric has been shown to sometimes better correlate with accuracy than using each metric individually \citep{detavernier2025comb}.

Lastly, since the strategies RS-D and RS-I are somewhat complementary of each other, devising a new technique that combines both of them is also a point of interest. Since this complementarity seems to be related to the presence or absence of a base model with superior predictive capacities than all others, one strategy to choose between using RS-D and RS-I would be to perform a hypothesis test. By performing multiple train-validation splits and obtaining the accuracies of the models for each split, one could test if the mean accuracy between the best performing model and all others is significantly different. If so, this could corroborate using RS-I instead of RS-D.

\section{Proof of Theorem 1}\label{sec:proof}
    Taking into account Equation~\eqref{eq:dcorfromdstarcor}, it clearly suffices to prove that $r_{d^*_{\mathrm{COR}}}(x)=r$, with $r=\frac{p(\hat{y}_1,x)}{p(\hat{y}_2,x)}$.

We first prove that $r_{d^*_{\mathrm{COR}}}(x)\leq r$, by constructing a measure $Q\in\mathbb{P}$ with density $q$ such that $d^*_{\mathrm{COR}}(P,Q)=r$ and $q(\hat{y}_2\vert x)\geq q(\hat{y}_1\vert x)$. 

Let
\begin{align*}
    &\pi_1:=P(Y=\hat{y}_1),~~~\pi_2:=P(Y=\hat{y}_2),\\
&\lambda_-:=\frac{\pi_1+\pi_2}{\pi_1+r\pi_2}\text{~~and~~}\lambda_+:=\frac{r\pi_1+r\pi_2}{\pi_1+r\pi_2}
\end{align*}
and consider the measurable function $L\colon\mathcal{Y}\times\mathcal{X}\to\mathbb{R}_{>0}$ defined as
$$L(y,x'):=\begin{cases}
\lambda_-&\text{if }y=\hat{y}_1,\\
\lambda_+&\text{if }y=\hat{y}_2,\\
1&\text{otherwise}.
\end{cases}$$
Then 
\begin{align*}\mathbb E_P[L] &= \lambda_+\pi_1+\lambda_-\pi_2+(1-\pi_1-\pi_2)\\
&=1+\pi_1(\lambda_+-1)+\pi_2(\lambda_--1)\\
&=1+\pi_1\frac{(r-1)\pi_2}{\pi_1+r\pi_2}-\pi_2\frac{(r-1)\pi_1}{\pi_1+r\pi_2}
=1.
\end{align*}
This implies that $Q$ defined by $\nicefrac{dQ}{dP}=L$ is a probability measure that is absolutely continuous w.r.t.\ $P$ and has density $q(y,x')=L(y,x')p(y,x')$. Moreover, since $\lambda_-\leq1\leq\lambda_+$, we have $\essinf L=\lambda_-$ and $\esssup L=\lambda_+$, so it follows from Equation~\eqref{eq:d_COR_simplified} that $$d^*_{\mathrm{COR}}(P,Q)=\frac{\esssup L}{\essinf L}=\frac{\lambda_+}{\lambda_-}=r.$$
Finally,
 $$\frac{q(\hat{y}_2\vert x)}{q(\hat{y}_1\vert x)}
 =\frac{q(\hat{y}_2,x)}{q(\hat{y}_1,x)}=\frac{\lambda_+p(\hat{y}_2,x)}{\lambda_-p(\hat{y}_1,x)}=\frac{r}{r}=1,$$
so $q(\hat{y}_2\vert x)\geq q(\hat{y}_1\vert x)$, which implies that the prediction $\hat{y}_1$ is not robust w.r.t.\ $\mathcal{P}_r$ for $d=d^*_{\mathrm{COR}}$ and thus that $r_{d^*_{\mathrm{COR}}}(x)\leq r$.

Next, we prove that $r_{d^*_{\mathrm{COR}}}(x)\geq r$. Let $Q\in\mathbb{P}$ be such that $d^*_{\mathrm{COR}}(P,Q)<r$, let $q$ be its density and $L=\nicefrac{q}{p}$ the likelihood ratio. Then $q(y,x)=L(y,x)p(y,x)$ and, due to Equation~\eqref{eq:d_COR_simplified}, $\esssup L/\essinf L<r$. For all $y\ne\hat{y}_1$, since $p(y,x)\leq p(\hat{y}_2,x)$, this implies that
\begin{align*}
\frac{q(\hat{y}_1\vert x)}{q(y\vert x)}
&=\frac{q(\hat{y}_1,x)}{q(y,x)}
=\frac{p(\hat{y}_1,x)}{p(y,x)}\frac{L(\hat{y}_1,x)}{L(y,x)}\\
&\geq\frac{p(\hat{y}_1,x)}{p(y,x)}\frac{\essinf L}{\esssup L}
>\frac{p(\hat{y}_1,x)}{p(y,x)}\frac{1}{r}\\
&\geq
\frac{p(\hat{y}_1,x)}{p(\hat{y}_2,x)}\frac{1}{r}=1.
\end{align*}
So $q(\hat{y}_1\vert x)>q(y\vert x)$ for all $y\ne\hat{y}_1$, which means that $Q$ predicts the same class $\hat{y}_1$ as $P$. Since this is true for every $Q\in\mathbb{P}$ such that $d^*_{\mathrm{COR}}(P,Q)<r$, we find that the prediction $\hat{y}_1$ is robust w.r.t.\ $\mathcal{P}_\delta$ for all $\delta<r$ and therefore that $r_{d^*_{\mathrm{COR}}}(x)\geq r$.

\begin{contributions} 
    The authors contributed equally to the development of the ideas and the writing of the paper. J.~De~Bock did the proof of \autoref{thm:rob} and R.~F.~L.~Lassance created the code and ran the experiments.

\end{contributions}

\begin{acknowledgements} 
    R.~F.~L.~Lassance: this work was carried out with the support of the Coordination for the Improvement of Higher Education Personnel – Brazil (CAPES) – Financing Code 001. J.~De~Bock: a BOF basic research funding grant of Ghent University was provided in the period that this research was carried out. The authors thank Adrián Detavernier and Rafael B. Stern for fruitful conversations and suggestions on how to make the contributions of robustness quantification more meaningful.

\end{acknowledgements}

\bibliography{uai2026-template}

@BOOK{augustin2014,
  title     = "Introduction to imprecise probabilities",
  editor    = "Augustin, Thomas and Coolen, Frank P. A. and de Cooman, Gert and
               Troffaes, Matthias C. M.",
  publisher = "Wiley-Blackwell",
  series    = "Wiley Series in Probability and Statistics",
  month     =  may,
  year      =  2014,
  address   = "Hoboken, NJ",
  language  = "en",
  doi       = "10.1002/9781118763117"
}

@article{breiman2001,
author = {Breiman, Leo},
title = {Random Forests},
year = {2001},
issue_date = {October 1 2001},
publisher = {Kluwer Academic Publishers},
address = {USA},
volume = {45},
number = {1},
issn = {0885-6125},
doi = {10.1023/A:1010933404324},
journal = {Mach. Learn.},
month = oct,
pages = {5–32},
numpages = {28}
}

@inproceedings{chen2016,
 author = {Chen, Tianqi and Guestrin, Carlos},
 title = {{XGBoost}: A Scalable Tree Boosting System},
 booktitle = {Proceedings of the 22nd ACM SIGKDD International Conference on Knowledge Discovery and Data Mining},
 series = {KDD '16},
 year = {2016},
 isbn = {978-1-4503-4232-2},
 location = {San Francisco, California, USA},
 pages = {785--794},
 numpages = {10},
 doi = {10.1145/2939672.2939785},
 acmid = {2939785},
 publisher = {ACM},
 address = {New York, NY, USA},
}

@InProceedings{correia2019,
author="Correia, Alvaro H. C.
and de Campos, Cassio P.",
editor="Ben Amor, Nahla
and Quost, Benjamin
and Theobald, Martin",
title="Towards Scalable and Robust Sum-Product Networks",
doi="10.1007/978-3-030-35514-2_31",
booktitle="Scalable Uncertainty Management",
year="2019",
publisher="Springer International Publishing",
address="Cham",
pages="409--422"
}

@misc{correia2020rob,
      title={Towards Robust Classification with Deep Generative Forests}, 
      author={Alvaro H. C. Correia and Robert Peharz and Cassio de Campos},
      year={2020},
      eprint={2007.05721},
      archivePrefix={arXiv},
      primaryClass={stat.ML},
      url={https://arxiv.org/abs/2007.05721}, 
}

@article{cruz2018,
title = {Dynamic classifier selection: Recent advances and perspectives},
journal = {Information Fusion},
volume = {41},
pages = {195-216},
year = {2018},
issn = {1566-2535},
doi = {10.1016/j.inffus.2017.09.010},
author = {Rafael M.O. Cruz and Robert Sabourin and George D.C. Cavalcanti}
}

@article{cruz2020,
    author  = {Rafael M. O. Cruz and Luiz G. Hafemann and Robert Sabourin and George D. C. Cavalcanti},
    title   = {DESlib: A Dynamic ensemble selection library in Python},
    journal = {Journal of Machine Learning Research},
    year    = {2020},
    volume  = {21},
    number  = {8},
    pages   = {1-5},
    url     = {http://jmlr.org/papers/v21/18-144.html}
}

@inproceedings{debock2014,
 author = {De Bock, Jasper and de Campos, Cassio P. and Antonucci, Alessandro},
 booktitle = {Advances in Neural Information Processing Systems},
 editor = {Z. Ghahramani and M. Welling and C. Cortes and N. Lawrence and K.Q. Weinberger},
 pages = {},
 publisher = {Curran Associates, Inc.},
 title = {Global Sensitivity Analysis for MAP Inference in Graphical Models},
 url = {https://proceedings.neurips.cc/paper_files/paper/2014/file/f9d3c99bd6cbf2d694266e7760ee1ed6-Paper.pdf},
 volume = {27},
 year = {2014}
}

@inproceedings{detavernier2025rob,
  author       = {Adri\'an Detavernier and Jasper {De Bock}},
  booktitle    = {{Proceedings of the Fourteenth International Symposium on Imprecise Probabilities: Theories and Applications}},
  issn         = {{2640-3498}},
  language     = {{eng}},
  location     = {{Bielefeld, Germany}},
  pages        = {{126--136}},
  title        = {{Robustness quantification : a new method for assessing the reliability of the predictions of a classifier}},
  url          = {{https://proceedings.mlr.press/v290/detavernier25a.html}},
  volume       = {{290}},
  year         = {{2025a}},
}

@misc{detavernier2025comb,
      title={Robustness and uncertainty: two complementary aspects of the reliability of the predictions of a classifier}, 
      author={Adri\'an Detavernier and Jasper {De Bock}},
      year={2025b},
      eprint={2512.15492},
      archivePrefix={arXiv},
      primaryClass={cs.LG},
      url={https://arxiv.org/abs/2512.15492}, 
}

@article{dumbgen2021,
  title = {Bounding distributional errors via density ratios},
  volume = {27},
  ISSN = {1350-7265},
  DOI = {10.3150/20-bej1256},
  number = {2},
  journal = {Bernoulli},
  publisher = {Bernoulli Society for Mathematical Statistics and Probability},
  author = {D\"{u}mbgen,  Lutz and Samworth,  Richard J. and Wellner,  Jon A.},
  year = {2021},
  month = may 
}

@Article{hullermeier2021,
author={H{\"u}llermeier, Eyke
and Waegeman, Willem},
title={Aleatoric and epistemic uncertainty in machine learning: an introduction to concepts and methods},
journal={Machine Learning},
year={2021},
month={Mar},
day={01},
volume={110},
number={3},
pages={457-506},
issn={1573-0565},
doi={10.1007/s10994-021-05946-3}
}

@Article{li2020,
AUTHOR = {Li, Meizhu and Huang, Shaoguang and De Bock, Jasper and de Cooman, Gert and Pižurica, Aleksandra},
TITLE = {A Robust Dynamic Classifier Selection Approach for Hyperspectral Images with Imprecise Label Information},
JOURNAL = {Sensors},
VOLUME = {20},
YEAR = {2020},
NUMBER = {18},
ARTICLE-NUMBER = {5262},
PubMedID = {32942592},
ISSN = {1424-8220},
DOI = {10.3390/s20185262}
}

@book{molnar2025,
  title={Interpretable Machine Learning},
  subtitle={A Guide for Making Black Box Models Explainable},
  author={Christoph Molnar},
  year={2025},
  edition={3},
  isbn={978-3-911578-03-5},
  publisher = {https://christophm.github.io/interpretable-ml-book/},
  url={https://christophm.github.io/interpretable-ml-book}
}

@article{montes2020,
author = {Ignacio Montes and Enrique Miranda and Sébastien Destercke},
title = {Unifying neighbourhood and distortion models: part I – new results on old models},
journal = {International Journal of General Systems},
volume = {49},
number = {6},
pages = {602--635},
year = {2020},
publisher = {Taylor \& Francis},
doi = {10.1080/03081079.2020.1778682}
}

@ARTICLE{muhammad2022,
  author={Muhammad, Awais and Bae, Sung-Ho},
  journal={IEEE Access}, 
  title={A Survey on Efficient Methods for Adversarial Robustness}, 
  year={2022},
  volume={10},
  number={},
  pages={118815-118830},
  doi={10.1109/ACCESS.2022.3216291}}

@InProceedings{nadeem2009,
  title = 	 {Accuracy-Rejection Curves (ARCs) for Comparing Classification Methods with a Reject Option},
  author = 	 {Nadeem, Malik Sajjad Ahmed and Zucker, Jean-Daniel and Hanczar, Blaise},
  booktitle = 	 {Proceedings of the third International Workshop on Machine Learning in Systems Biology},
  pages = 	 {65--81},
  year = 	 {2009},
  editor = 	 {Džeroski, Sašo and Guerts, Pierre and Rousu, Juho},
  volume = 	 {8},
  series = 	 {Proceedings of Machine Learning Research},
  address = 	 {Ljubljana, Slovenia},
  month = 	 {05--06 Sep},
  publisher =    {PMLR},
  url = 	 {https://proceedings.mlr.press/v8/nadeem10a.html}
}

@book{oneil2016,
author = {O'Neil, Cathy},
title = {Weapons of Math Destruction: How Big Data Increases Inequality and Threatens Democracy},
year = {2016},
isbn = {0553418815},
publisher = {Crown Publishing Group},
address = {USA}
}

@article{openml,
author = {Vanschoren, Joaquin and van Rijn, Jan N. and Bischl, Bernd and Torgo, Luis},
title = {{OpenML}: networked science in machine learning},
year = {2014},
issue_date = {December 2013},
publisher = {Association for Computing Machinery},
address = {New York, NY, USA},
volume = {15},
number = {2},
issn = {1931-0145},
doi = {10.1145/2641190.2641198},
journal = {SIGKDD Explor. Newsl.},
month = jun,
pages = {49–60},
numpages = {12}
}

@article{romano2021,
  title={{PMLB} v1.0: an open source dataset collection for benchmarking machine learning methods},
  author={Romano, Joseph D. and Le, Trang T. and La Cava, William and Gregg, John T. and Goldberg, Daniel J. and Chakraborty, Praneel and Ray, Natasha L. and Himmelstein, Daniel and Fu, Weixuan and Moore, Jason H.},
  journal={arXiv preprint arXiv:2012.00058v2},
  year={2021},
  url={https://arxiv.org/abs/2012.00058}
}

@Article{rudin2019,
author={Rudin, Cynthia},
title={Stop explaining black box machine learning models for high stakes decisions and use interpretable models instead},
journal={Nature Machine Intelligence},
year={2019},
month={May},
day={01},
volume={1},
number={5},
pages={206-215},
issn={2522-5839},
doi={10.1038/s42256-019-0048-x}
}

@article{scikit,
  author  = {Fabian Pedregosa and Ga{{\"e}}l Varoquaux and Alexandre Gramfort and Vincent Michel and Bertrand Thirion and Olivier Grisel and Mathieu Blondel and Peter Prettenhofer and Ron Weiss and Vincent Dubourg and Jake Vanderplas and Alexandre Passos and David Cournapeau and Matthieu Brucher and Matthieu Perrot and {{\'E}}douard Duchesnay},
  title   = {Scikit-learn: Machine Learning in Python},
  journal = {Journal of Machine Learning Research},
  year    = {2011},
  volume  = {12},
  number  = {85},
  pages   = {2825--2830},
  url     = {http://jmlr.org/papers/v12/pedregosa11a.html}
}

@misc{uci,
  author = {Kelly, Markelle and Longjohn, Rachel and Nottingham, Kolby},
  title = {{UCI} Machine Learning Repository},
  URL = {https://archive.ics.uci.edu},
  year = {2023}
}

@BOOK{Walley1991,
  title = {Statistical Reasoning with Imprecise Probabilities},
  author = {Peter Walley},
  publisher = {Chapman and Hall},
  year = {1991},
  address = {London},
  series = {Monographs on Statistics and Applied Probability},
  volume = {42}
}

\newpage

\onecolumn

\title{Robustness Quantification for Discriminative Models\\(Supplementary Material)}
\maketitle

\appendix

\begin{table}[!htb]
    \centering
    \caption{Mean accuracy of each method under label corruption ($\rho = 0.1$).\\
    \small  Best accuracy in bold, second best underlined.}\label{tab:mean_acc_noise10}
    \fontsize{8.7pt}{9pt}\selectfont
    \begin{tabular}{l|ccc|ccccccc}
        Label &GB&RF&XGB            & SB & MCB & KNORA-U & KNORA-E & META-DES & RS-D & RS-I\\
        \hline
        $D_{1}$ &0.98068 & 0.98261 & 0.98647& 0.98841 & 0.98583 & 0.98937 & \underline{0.98969} & \textbf{0.99002} & 0.98776 & \underline{0.98969}\\
        $D_{2}$ &0.91351 & 0.90522 & 0.91446& 0.91385 & 0.91166 & 0.91145 & 0.9113  & 0.91284 & \underline{0.91423} & \textbf{0.91429}\\
        $D_{3}$ &0.74713 & 0.76954 & 0.7546& 0.75402 & \textbf{0.76609} & 0.76149 & 0.7569  & 0.76207 & \underline{0.76379} & 0.75575\\
        $D_{4}$ &0.85802 & 0.89204 & 0.87812& 0.89204 & 0.88107 & 0.88344 & 0.88445 & 0.88689 & \textbf{0.89383} & \underline{0.89219}\\
        $D_{5}$ &0.58817 & 0.65092 & 0.63927& 0.64975 & 0.63437 & 0.6305  & 0.64076 & 0.63711 & \textbf{0.65245} & \underline{0.65074}\\
        $D_{6}$ &0.87064 & 0.87786 & 0.87342& 0.87648 & 0.87416 & 0.87489 & 0.8747  & 0.87564 & \textbf{0.87741} & \underline{0.87687}\\
        $D_{7}$ &0.98803 & 0.9676  & 0.98958& 0.98942 & 0.98551 & 0.99088 & 0.99105 & \textbf{0.99162} & \underline{0.99129} & 0.99064\\
        $D_{8}$ &0.94409 & 0.92065 & 0.96119& \underline{0.96061} & 0.95293 & 0.95793 & 0.95812 & 0.95869 & 0.95985 & \textbf{0.96081}\\
        $D_{9}$ &0.75669 & 0.76622 & 0.7612& 0.7614  & 0.76221 & 0.76241 & 0.76371 & \underline{0.76411} & \textbf{0.76571} & 0.76361\\
        $D_{10}$&0.95956 & 0.86772 & 0.97869& \textbf{0.97869} & 0.96053 & 0.97103 & 0.97256 & 0.97119 & 0.97635 & \underline{0.97797}\\
        $D_{11}$&0.9613  & 0.96874 & 0.95842& \textbf{0.9679}  & 0.96202 & 0.96754 & 0.9673  & 0.96748 & \underline{0.96778} & 0.9676\\
        $D_{12}$&0.82685 & 0.91991 & 0.87919& \textbf{0.92081} & 0.8868  & 0.89306 & 0.89575 & 0.89083 & 0.90917 & \underline{0.9132}\\
        $D_{13}$&0.84412 & 0.85237 & 0.84874& 0.84891 & 0.85166 & \underline{0.85184} & \textbf{0.85344} & 0.85122 & 0.849   & 0.84971\\
        $D_{14}$&0.84669 & 0.85211 & 0.85069& 0.84927 & 0.84927 & 0.85149 & 0.85246 & \underline{0.85477} & \textbf{0.85584} & 0.85078\\
        $D_{15}$&0.92248 & 0.94574 & 0.93721& 0.94186 & 0.94109 & 0.94186 & \underline{0.94419} & 0.93876 & \textbf{0.94496} & 0.94264\\
        \hline
        Beats SB &0/15&8/15&6/15& - & 3/15 & 6/15 & 7/15 & 6/15 & 10/15 & 12/15\\
        \hline
    \end{tabular}
\end{table}

\begin{table}[!htb]
    \centering
    \caption{Mean accuracy of each method under label corruption ($\rho = 0.2$).\\
    \small  Best accuracy in bold, second best underlined.}\label{tab:mean_acc_noise20}
    \fontsize{8.7pt}{9pt}\selectfont
    \begin{tabular}{l|ccc|ccccccc}
        Label &GB&RF&XGB            &SB & MCB & KNORA-U & KNORA-E & META-DES & RS-D & RS-I\\
        \hline
        $D_{1}$ &0.94847 & 0.97359 & 0.97262& 0.97327 & 0.96844 & \underline{0.97874} & \textbf{0.97939} & 0.97617 & 0.97649 & 0.9752\\
        $D_{2}$ &0.91295 & 0.90548 & 0.91228& 0.91313 & 0.91017 & 0.9105  & 0.91051 & 0.91242 & \textbf{0.91338} & \underline{0.91314}\\
        $D_{3}$ &0.71322 & 0.75287 & 0.73908& 0.74195 & 0.73046 & \textbf{0.74483} & \underline{0.74425} & 0.7431  & 0.73851 & 0.73851\\
        $D_{4}$ &0.84607 & 0.85835 & 0.8595& 0.85955 & 0.859   & 0.86329 & \underline{0.86388} & 0.86241 & \textbf{0.86998} & 0.86176\\
        $D_{5}$ &0.56644 & 0.64197 & 0.62231& 0.63941 & 0.61831 & 0.61165 & 0.62434 & 0.62092 & \textbf{0.64462} & \underline{0.64017}\\
        $D_{6}$ &0.86403 & 0.8665  & 0.86578& 0.86529 & 0.86508 & 0.86802 & 0.86847 & \textbf{0.8687}  & \underline{0.86858} & 0.86559\\
        $D_{7}$ &0.98266 & 0.96443 & 0.9825& 0.98063 & 0.98128 & 0.98584 & \underline{0.98689} & \textbf{0.98746} & 0.98665 & 0.98396\\
        $D_{8}$ &0.92277 & 0.92027 & 0.9414& 0.93929 & 0.93372 & 0.94159 & \textbf{0.94467} & \underline{0.94428} & 0.94409 & 0.94121\\
        $D_{9}$ &0.75028 & 0.75228 & 0.74657& 0.74937 & 0.74817 & 0.75388 & \textbf{0.75589} & 0.75409 & \underline{0.75519} & 0.75258\\
        $D_{10}$&0.94649 & 0.86489 & 0.96594& \underline{0.96594} & 0.94496 & 0.95642 & 0.96287 & 0.95932 & \underline{0.96594} & \textbf{0.96626}\\
        $D_{11}$&0.95014 & 0.96538 & 0.94287& \textbf{0.96538} & 0.95476 & 0.95884 & 0.9586 & 0.9589 & 0.96244 & \underline{0.96436}\\
        $D_{12}$&0.74989 & 0.8783  & 0.82998& \textbf{0.8783}  & 0.83669 & 0.83982 & 0.84743 & 0.85101 & 0.87293 & \underline{0.8774}\\
        $D_{13}$&0.83604 & 0.85308 & 0.83356& 0.8474  & 0.8419  & 0.84359 & 0.84634 & 0.84634 & \underline{0.84758} & \textbf{0.84838}\\
        $D_{14}$&0.83595 & 0.85362 & 0.83693& \underline{0.85175} & 0.8435  & 0.8466  & 0.8498  & 0.849   & \textbf{0.85255} & 0.85166\\
        $D_{15}$&0.88915 & 0.92946 & 0.90775& 0.92016 & 0.9093  & 0.92481 & \underline{0.92558} & 0.92403 & \textbf{0.92946} & 0.92403\\
        \hline
        Beats SB&2/15&8/15&5/15& - & 1/15 & 8/15 & 9/15 & 8/15 & 11/15 & 11/15\\
        \hline
    \end{tabular}
\end{table}

\end{document}